\documentclass[a4paper,fleqn]{cas-dc}

\usepackage[authoryear,longnamesfirst]{natbib}
\definecolor{lightred}{RGB}{255,185,185}
\definecolor{lightorange}{RGB}{255,210,150}
\definecolor{lightyellow}{RGB}{255,248,200}
\usepackage{pifont}
\newcommand{\cmark}{\ding{51}}
\newcommand{\xmark}{\ding{55}}
\usepackage{graphicx}
\usepackage{float} 
\usepackage{enumitem}
\usepackage{bm}

\def\tsc#1{\csdef{#1}{\textsc{\lowercase{#1}}\xspace}}
\tsc{WGM}
\tsc{QE}
\begin{document}
\let\WriteBookmarks\relax
\def\floatpagepagefraction{1}
\def\textpagefraction{.001}

\shorttitle{SWAGSplatting}    

\shortauthors{Jiang et al.}  

\title [mode = title]{Semantic-guided Gaussian Splatting for High-Fidelity Underwater Scene Reconstruction}  




\author[1]{Zhuodong Jiang}[orcid=0009-0002-5000-2752]

\cormark[1]

\fnmark[1]

\ead{zhuodong.jiang@bristol.ac.uk}


\credit{Writing - review \& editing, Writing - original draft, Visualization, Validation, Methodology, Data curation, Conceptualization}

\affiliation[1]{organization={School of Computer Science, University of Bristol},
            addressline={Bristol}, 
            % city={},
%          citysep={}, % Uncomment if no comma needed between city and postcode
            country={UK}}

\author[1]{Haoran Wang}

% Footnote of the second author
\fnmark[1]

% Email id of the second author
\ead{yp22378@bristol.ac.uk}


\credit{Writing - review \& editing, Writing - original draft, Visualization, Validation, Methodology, Conceptualization}

\affiliation[2]{organization={Submerged Resources Centre},
            addressline={Denver}, 
            % city={},
%          citysep={}, % Uncomment if no comma needed between city and postcode
            state={CO},
            country={USA}}

\author[1]{Guoxi Huang}
\ead{guoxi.huang@bristol.ac.uk}
\credit{Writing - review \& editing, Methodology}

\author[2]{Brett Seymour}
\ead{Brett_Seymour@nps.gov}
\credit{Data curation, Resources}

\author[1]{Nantheera Anantrasirichai}[orcid=0000-0002-2122-5781]
\ead{n.anantrasirichai@bristol.ac.uk}
\credit{Writing - review \& editing, Methodology, Conceptualization, Supervision, Resources, Project administration, Funding acquisition}
\cortext[1]{Corresponding author}

\fntext[1]{Equal contribution}


\begin{abstract}
Accurate 3D reconstruction in degraded imaging conditions remains a key challenge in photogrammetry and neural rendering. In underwater environments, spatially varying visibility caused by scattering, attenuation, and sparse observations leads to highly non-uniform information quality. Existing 3D Gaussian Splatting (3DGS) methods typically optimize primitives based on photometric signals alone, resulting in imbalanced representation, with overfitting in well-observed regions and insufficient reconstruction in degraded areas.
In this paper, we propose SWAGSplatting (Semantic-guided Water-scene Augmented Gaussian Splatting), a multimodal framework that integrates semantic priors into 3DGS for robust, high-fidelity underwater reconstruction. Each Gaussian primitive is augmented with a learnable semantic feature, supervised by CLIP-based embeddings derived from region-level cues. A semantic consistency loss is introduced to align geometric reconstruction with high-level semantics, improving structural coherence and preserving salient object boundaries under challenging conditions. 
Furthermore, we propose an adaptive Gaussian primitive reallocation strategy that redistributes representation capacity based on both primitive importance and reconstruction error, mitigating the imbalance introduced by conventional densification. This enables more effective modeling of low-visibility regions without increasing computational cost.
Extensive experiments on real-world datasets, including SeaThru-NeRF, Submerged3D, and S-UW, demonstrate that the proposed method consistently outperforms state-of-the-art approaches in terms of average PSNR, SSIM, and LPIPS. The results validate the effectiveness of integrating semantic priors for high-fidelity underwater scene reconstruction. Code is available at \url{https://github.com/theflash987/SWAGSplatting}.
\end{abstract}


% \begin{highlights}
% \item 
% \item 
% \item 
% \end{highlights}

%\nocite{*}

% Keywords

\maketitle

% Main text
\section{Introduction}\label{sec:intro}
Underwater exploration supports a wide range of applications, including robotics, archaeology, and marine ecology, all of which rely on accurate 3D reconstruction for visualization, interpretation, and navigation. Active sensing modalities such as sonar and LiDAR provide direct range measurements, but their use underwater is limited by resolution, cost, and operational complexity \citep{s21237849}. Acoustic sensors enable long-range perception but typically produce low-resolution reconstructions, while underwater LiDAR suffers from strong attenuation and backscatter that restricts effective range. In contrast, image-based methods rely on passive sensing and widely available camera systems, offering lower cost and higher spatial resolution for detailed reconstruction. These advantages make them particularly attractive; however, underwater environments present unique challenges, such as limited visibility, color distortion, sparse viewpoints, marine snows, and noise caused by light attenuation and scattering. Capturing videos in deep underwater settings is especially difficult, as low light conditions require longer exposure times or higher ISO settings, both of which introduce significant motion blur and sensor noise. Still image capture is often limited and unevenly distributed, further compounding the difficulty of reliable 3D reconstruction in such conditions.

Photogrammetry-based methods have long formed the foundation of 3D scene reconstruction, including in underwater environments~\citep{ELNASHEF2023223,ZHONG2025779}. These approaches recover geometry from multi-view image correspondences and camera calibration, typically through pipelines such as Structure-from-Motion (SfM) and Multi-View Stereo (MVS), which estimate camera poses and dense surface geometry under favorable imaging conditions. However, their performance degrades in challenging scenarios characterized by low texture, scattering, and non-uniform illumination, where reliable feature matching and depth estimation become difficult~\citep{SONG2024197}. In addition, such pipelines can be computationally demanding, limiting their applicability in large-scale or time-sensitive settings.

Recent advances in neural rendering have introduced alternative paradigms for 3D reconstruction through novel view synthesis (NVS). Neural Radiance Fields (NeRF) ~\citep{mildenhall2021nerf} achieve high-quality reconstructions by modeling continuous volumetric representations, but require dense input views and incur significant computational cost. More recently, 3D Gaussian Splatting (3DGS)~\citep{kerbl:3Dgaussians:2023} has emerged as an efficient explicit representation, enabling faster optimization and real-time rendering. Despite these advantages, most NeRF- and 3DGS-based approaches assume clear imaging conditions and exhibit substantial performance degradation in turbid underwater environments. Several methods have been proposed that integrate underwater image degradation models~\citep{levy2023seathru,wang2025uw,li2024watersplatting,jiang2025rusplatting}, showing promising improvements in 3D reconstruction, yet a critical limitation persists: these methods treat all scene regions uniformly during optimization.

In underwater settings, visual information is spatially non-uniform. Foreground objects typically provide stronger and more reliable signals than background regions affected by scattering and attenuation. Without explicit guidance, standard optimization processes prioritize well-observed regions, diluting contributions from semantically important or degraded areas. This results in reconstructions where salient structures appear blurred or geometrically inconsistent. A related limitation arises from gradient-driven Gaussian densification, which allocates primitives according to photometric signal strength. Under degraded imaging conditions, this results in an imbalanced primitive distribution, in which well-observed regions accumulate redundant Gaussians, while challenging areas remain under-represented, ultimately limiting reconstruction fidelity.

To address these limitations, we introduce a semantic-augmented Gaussian splatting framework for underwater 3D reconstruction, SWAGSplatting (\textbf{S}emantic-guided \textbf{W}ater-scene \textbf{A}ugmented \textbf{G}aussian \textbf{S}platting). The proposed approach integrates semantic priors with physically grounded modeling to improve reconstruction under challenging imaging conditions. Specifically, semantic cues derived from vision–language models, such as CLIP~\citep{radford2021learning}, are used to provide high-level contextual information that complements photometric supervision. CLIP learns transferable visual–semantic correspondences from large-scale image–text data, enabling generalization across domains without task-specific training.
We incorporate this information through a semantic consistency loss, which acts as an auxiliary regularization term to align per-Gaussian semantic features with region-level semantic embeddings. Importantly, this regularization operates in feature space and does not directly modify geometric parameters, such as position or opacity. As a result, the proposed formulation enhances structural coherence and object-level consistency while preserving the underlying geometric representation.

\begin{figure}
\centering%% For centre alignment of image.
\includegraphics[width=1\columnwidth]{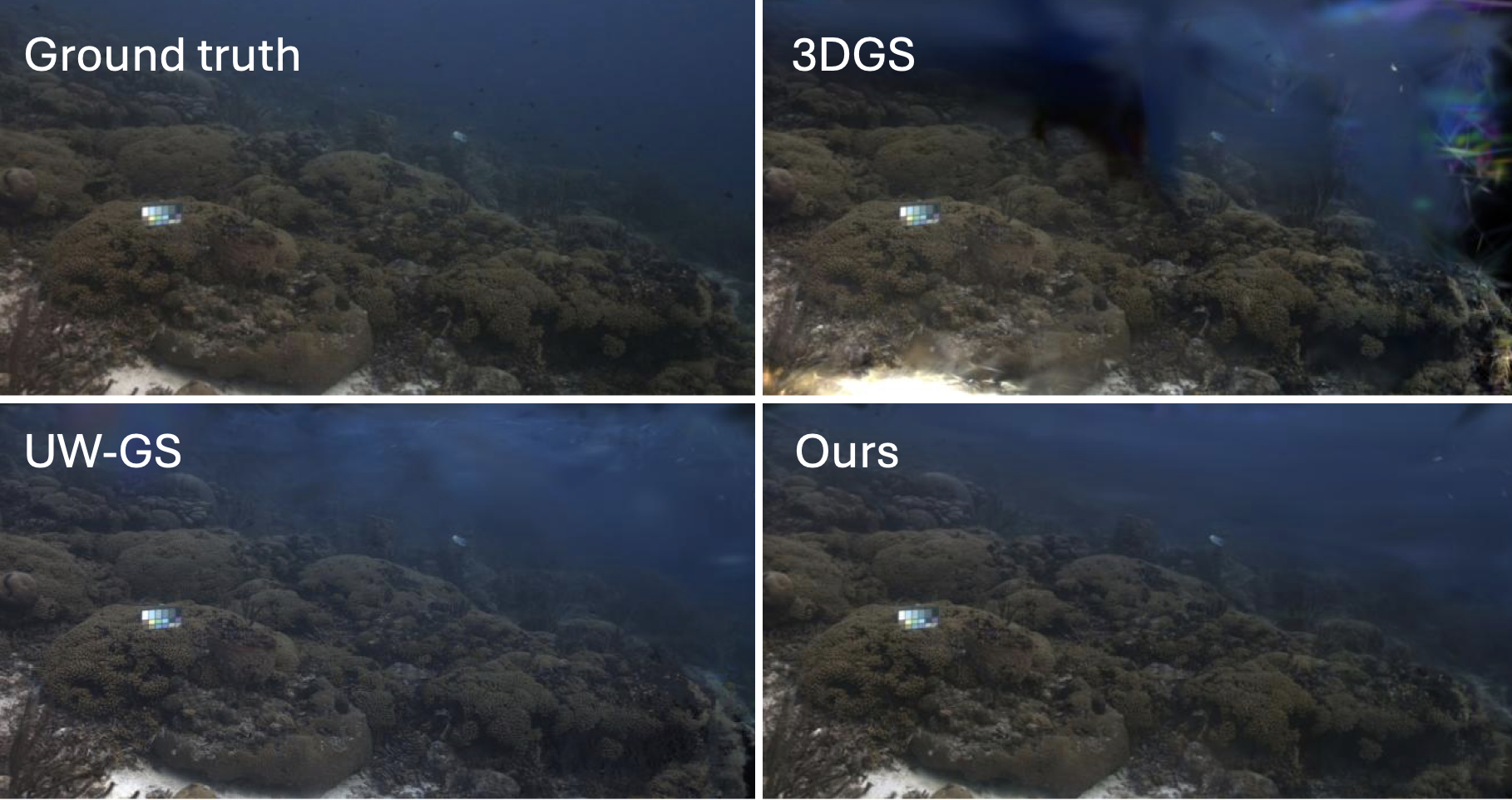}
\caption{Reconstruction performance comparison among 3DGS~\citep{kerbl:3Dgaussians:2023}, UW-GS~\citep{wang2025uw}, and the proposed SWAGSplatting (Ours). Relative to the 3DGS and UW-GS methods, SWAGSplatting markedly suppresses artifact generation and yields more faithful rendering results.}\label{fig:teaser}
\end{figure}

In addition, standard 3DGS relies on gradient-driven densification, which allocates primitives according to photometric signal strength. Under degraded imaging conditions, this leads to an imbalanced distribution of Gaussians, where well-observed regions dominate the optimization process while low-visibility areas remain under-represented. As a result, representation capacity is inefficiently utilized, with redundant primitives in high-signal regions and insufficient coverage in challenging areas, ultimately limiting reconstruction quality.
To address this limitation, we propose an adaptive Gaussian allocation strategy that redistributes representation capacity based on reconstruction difficulty, enabling more balanced spatial coverage and improved reconstruction fidelity as shown in \autoref{fig:teaser}.

The main contributions of this work are as follows:
\begin{itemize}[leftmargin=*]
    \item We propose a \textbf{semantic-guided 3D Gaussian Splatting framework} for underwater scene reconstruction, in which each Gaussian primitive is augmented with a learnable semantic feature derived from vision-language embeddings, enabling object-aware and semantically consistent reconstruction.
    \item We introduce a \textbf{semantic-guided loss} that enforces alignment between semantic and geometric representations, improving structural coherence and preserving salient object boundaries under degraded imaging conditions.
    \item \textbf{Adaptive Gaussian primitive reallocation} is proposed to balance the distribution of the Gaussian point cloud by reallocating low-importance primitives to high-error regions, thereby enhancing NVS quality.
    \item We introduce a \textbf{stage-wise optimization strategy}, a coarse-to-fine training scheme that enhances stability and visual quality via late-stage parameter freezing and $\ell_2$ fine-tuning.
\end{itemize}
We conduct a comprehensive evaluation on the SeaThru-NeRF~\citep{levy2023seathru}, Submerged3D~\citep{jiang2025rusplatting} and S-UW~\citep{wang2025uw} datasets, demonstrating an improvement of up to 3.48 dB in PSNR and consistent gains in SSIM and LPIPS over state-of-the-art baselines. 

% Use the style of numbering in square brackets.
% If nothing is used, default style will be taken.
%\begin{enumerate}[a)]
%\item 
%\item 
%\item 
%\end{enumerate}  

%\begin{itemize}
%\item 
%\item 
%\item 
%\end{itemize}  

%\begin{description}
%\item[]
%\item[] 
%\item[] 
%\end{description}  
\section{Related work}

Prior to the rise of neural rendering, underwater 3D reconstruction relied primarily on photogrammetric pipelines built on Structure-from-Motion (SfM) and Multi-View Stereo (MVS). Systems such as COLMAP~\citep{schonberger2016structure} extract sparse point clouds from feature correspondences across images and then densify them through stereo matching, forming the backbone of many underwater survey workflows~\citep{bryson2016true} and commercial software~\citep{Burns:compare:2017}. However, these methods assume Lambertian surfaces and consistent illumination, conditions that are rarely met underwater, where scattering, backscatter, and wavelength-dependent attenuation corrupt both feature detection and stereo matching. Preprocessing steps such as histogram equalization, dark channel prior dehazing, or Sea-thru color correction~\citep{akkaynak2019sea} are commonly applied to improve input quality, yet they cannot fully compensate for severe degradation, particularly in deep or turbid water. Acoustic sensing offers a complementary approach. This includes multibeam sonar and structured light systems that can recover geometry independently of optical conditions, but produce sparse or textureless reconstructions \citep{jmse11050949}.

More recent classical approaches have explored stereo vision and simultaneous localization and mapping (SLAM) for real-time underwater reconstruction. Stereo-based systems leverage epipolar geometry to recover dense depth maps, with underwater-specific calibration models accounting for refractive distortion at the housing interface~\citep{5770266, SHE2022525}. Visual SLAM systems adapted for underwater use, such as those integrating inertial measurements or sonar constraints, enable incremental map building during robot traversal~\citep{s25113258}. While these methods are well-suited to online navigation tasks, they typically produce lower-fidelity reconstructions than offline neural rendering pipelines and struggle with the appearance inconsistency introduced by dynamic lighting and medium scattering. The limitations of both photogrammetric and SLAM-based approaches in handling underwater optical degradation have driven the community towards learning-based methods, beginning with NeRF and more recently 3DGS.

\subsection{NeRF-based Underwater Scene Reconstruction}
The extension of NeRF~\citep{mildenhall2021nerf} to scattering media has attracted increasing attention, driven by the need to decouple scene geometry from medium-induced degradation. Early work such as ScatterNeRF~\citep{ramazzina2023scatternerf} explicitly models volumetric attenuation separately from object geometry, providing a physically grounded formulation for rendering in participating media.

Building on the revised underwater image formation model of~\citep{akkaynak2018revised}, SeaThru-NeRF ~\citep{levy2023seathru} introduces dedicated network components to model transmittance and medium color independently, enabling improved disentanglement of scene appearance from water effects. Subsequent work has focused on enhancing reconstruction quality under this framework. For example, SP-SeaNeRF~\citep{CHEN2024104025} incorporates learnable illumination embeddings to improve sharpness and detail preservation. Another line of research addresses color degradation through physically motivated models. WaterNeRF~\citep{Sethuraman:WaterNeRF:2023} combines light transport modeling with optimal transport for color correction, while WaterHE-NeRF~\citep{ZHOU2025102770} adopts a Retinex-based formulation to compensate for wavelength-dependent attenuation. These approaches emphasize the importance of accurate color modeling for perceptually consistent reconstruction.

Handling dynamic elements remains a key challenge in underwater environments. UWNeRF~\citep{tang2024neural} separates static and dynamic regions to improve reconstruction stability, although it relies on accurate masking. AquaNeRF~\citep{gough2025aquanerf} mitigates dynamic artifacts using a single-surface-per-ray formulation with Gaussian-weighted transmittance, reducing the dependence on explicit segmentation.

Despite these advances, NeRF-based methods remain constrained by their reliance on dense input views, high computational cost, and limited rendering efficiency. These limitations motivate the adoption of more efficient explicit representations, such as 3DGS, which we discuss next.

\subsection{3DGS-based Underwater Scene Reconstruction}

Recent advances in 3D Gaussian Splatting (3DGS)~\citep{kerbl:3Dgaussians:2023} have enabled efficient and high-quality scene reconstruction through explicit point-based representations. Compared to NeRF-based methods, 3DGS offers faster training and real-time rendering, making it attractive for large-scale and interactive applications. However, directly applying 3DGS to underwater environments remains challenging due to scattering, attenuation, and dynamic disturbances \citep{huang:Visual:2025}.

Early adaptations of 3DGS to underwater scenes focus on incorporating physical models of light propagation. UW-GS~\citep{wang2025uw} introduces a physics-aware density control mechanism, pseudo-depth supervision and motion masking to suppress the effects of scattering and dynamic distractors. WaterSplatting~\citep{li2024watersplatting} improves realism by explicitly modeling object and medium transmittance, enabling enhanced visual quality under varying water conditions. Aquatic-GS~\citep{liu2024aquatic} combines implicit water representations with explicit Gaussian primitives to achieve physically consistent reconstructions.

Subsequent research has explored a range of strategies to improve the robustness of neural rendering methods in underwater environments, where visibility degradation and illumination variability are pervasive. For example, SeaSplat~\citep{yang2025seasplat} incorporates an underwater image formation model into the 3DGS framework to disentangle scene radiance from medium effects, thereby enhancing both color fidelity and geometric reconstruction. However, this method is primarily tailored to static scenes and may be less effective in more complex or dynamic environments. RecGS~\citep{zhang2025recgs} employs a recurrent optimization strategy that progressively refines the reconstruction to suppress water-induced artifacts, leading to improved stability under degraded visual conditions. RUSplatting~\citep{jiang2025rusplatting} further introduces an uncertainty-aware modeling mechanism to estimate the reliability of observations and reduce the impact of noisy or corrupted inputs during optimization. Meanwhile, R-Splatting~\citep{huang2025fromrestoration} integrates pretrained image enhancement networks to alleviate illumination degradation and color distortion, which are especially pronounced in shallow-water scenes.

Another direction focuses on structural decomposition and representation design. AtlantisGS~\citep{Yi:AtlantisGS:2025} separates foreground objects from the background medium and allocates additional primitives to salient regions, improving reconstruction quality in sparse-view settings. These approaches highlight the importance of adapting Gaussian representations to account for scene structure and environmental effects.

Existing approaches largely focus on low-level image restoration or observation weighting, and often overlook the role of higher-level structural or semantic cues that could further stabilize reconstruction under spatially varying visibility. This limitation motivates the development of methods that incorporate additional priors to guide geometry optimization in challenging underwater environments.

\section{Preliminaries}

\subsection{3D Gaussian Splatting (3DGS)}

3D Gaussian Splatting (3DGS)~\citep{kerbl:3Dgaussians:2023} represents a scene as a set of explicit primitives in the form of anisotropic 3D Gaussians. Each Gaussian $G_i$ is parameterised by its mean position $\bm{\mu}_i \in \mathbb{R}^3$, covariance matrix $\Sigma_i \in \mathbb{R}^{3 \times 3}$, opacity $\alpha_i$ (or density contribution), and view-dependent color $c_i$. The covariance matrix encodes the spatial extent and orientation of each primitive, allowing anisotropic Gaussians to approximate local surface geometry more efficiently than isotropic representations.

During rendering, each Gaussian is projected from 3D space onto the 2D image plane using the camera projection model. Let $\bm{x}$ denote a pixel location in image space. The contribution of a projected Gaussian to $\bm{x}$ is modelled as a 2D Gaussian distribution:
\begin{equation}
G_i(\bm{x}) = \exp\left( -\frac{1}{2} (\bm{x} - \bm{\mu}_i)^\top \Sigma_i^{-1} (\bm{x} - \bm{\mu}_i) \right),
\end{equation}
where $\bm{\mu}_i$ and $\Sigma_i$ denote the projected mean and covariance in screen space. This formulation enables efficient rasterization by evaluating the influence of each Gaussian in image space. To facilitate a differentiable optimization, the covariance matrix $\Sigma_i$ is factorized into a rotation matrix $R_i \in \mathbb{R}^{3 \times 3}$ and a scaling matrix $S_i \in \mathbb{R}^{3 \times 3}$ in the 3D Gaussian primitive.
\begin{equation}
\Sigma_i = R_i S_i S_i^{T} R_i^{T}.
\end{equation}

To account for visibility and occlusion, Gaussians are composited using front-to-back alpha blending. The final color at pixel $\bm{x}$ is given by:
\begin{equation}
C(\bm{x}) = \sum_{i=1}^{N} \alpha_i \, c_i(\mathbf{v}) \prod_{j=1}^{i-1}(1 - \alpha_j),
\end{equation}
where $\mathbf{v}$ denotes the viewing direction, and $c_i(\mathbf{v})$ is typically represented using spherical harmonics (SH) to model view-dependent appearance. The multiplicative term accounts for accumulated transmittance along the ray, ensuring correct visibility ordering.

The parameters of the Gaussians, including position, covariance, color, and opacity, are optimized through gradient-based minimization of a photometric reconstruction loss between rendered and observed images. To improve coverage of the scene, 3DGS employs an adaptive densification strategy that introduces additional Gaussians in regions with high reconstruction gradients \citep{huang2025decomposing}. While effective in well-observed areas, such gradient-driven allocation can lead to an imbalanced distribution of primitives under non-uniform imaging conditions \citep{wang2026prune}, a limitation that we address in this work.

% -------------------------------------------------------

% -------------------------------------------------------
\subsection{Underwater Image Formation}

Underwater image formation is determined by the interaction between light and the surrounding medium, leading to significantly stronger degradation than in in-air imaging. In particular, light propagation is affected by wavelength-dependent absorption and scattering, which result in reduced contrast, color distortion, and spatially varying visibility~\citep{tang2024neural}. These effects pose significant challenges for image-based 3D reconstruction, as they violate the common assumptions of consistent appearance and reliable feature correspondences.

Following the widely adopted underwater imaging model ~\citep{akkaynak2018revised}, the observed color $I_c$ at a camera pixel can be expressed as the combination of attenuated direct signal and backscattered light:
\begin{equation}
I_c = J \cdot T^D + B^{\infty} \cdot (1 - T^B),
\end{equation}
where $J$ denotes the true scene radiance, and $B^{\infty}$ represents the background light contributed by the medium. The transmission terms $T^D$ and $T^B$ correspond to the direct and backscattered components, respectively, and are modelled as:
\begin{equation}
\begin{aligned}
T^D &= \exp(-\beta^d \cdot z), \\
T^B &= \exp(-\beta^b \cdot z),
\end{aligned}
\end{equation}
where $\beta^d$ and $\beta^b$ are attenuation coefficients and $z$ denotes the distance between the scene point and the camera.

The direct transmission term $T^D$ models the exponential decay of scene radiance due to absorption and out-scattering along the line of sight, leading to progressive loss of contrast with increasing depth. The backscattered component $T^B$ accounts for light scattered by suspended particles toward the camera, introducing a veiling effect that reduces image clarity and obscures fine structural details \citep{9105267}. These effects are further compounded by wavelength-dependent attenuation, where shorter wavelengths (e.g., blue) propagate further than longer wavelengths (e.g., red), resulting in characteristic color shifts in underwater imagery.

From a reconstruction perspective, these physical processes lead to several challenges. First, attenuation reduces the signal-to-noise ratio in distant regions, making feature detection and matching unreliable. Second, backscatter introduces spatially varying haze that degrades geometric consistency across views. Third, the dependence on depth $z$ creates non-uniform visibility across the scene, such that different regions are observed with varying levels of fidelity. As a result, standard reconstruction methods tend to favor well-observed regions while under-representing degraded areas. In the context of 3DGS, incorporating such physical considerations is essential for achieving balanced representation and robust reconstruction under underwater conditions.

% ============================================================

\begin{figure*}[pos=bt]
  \centering
  \includegraphics[width=\textwidth]{ 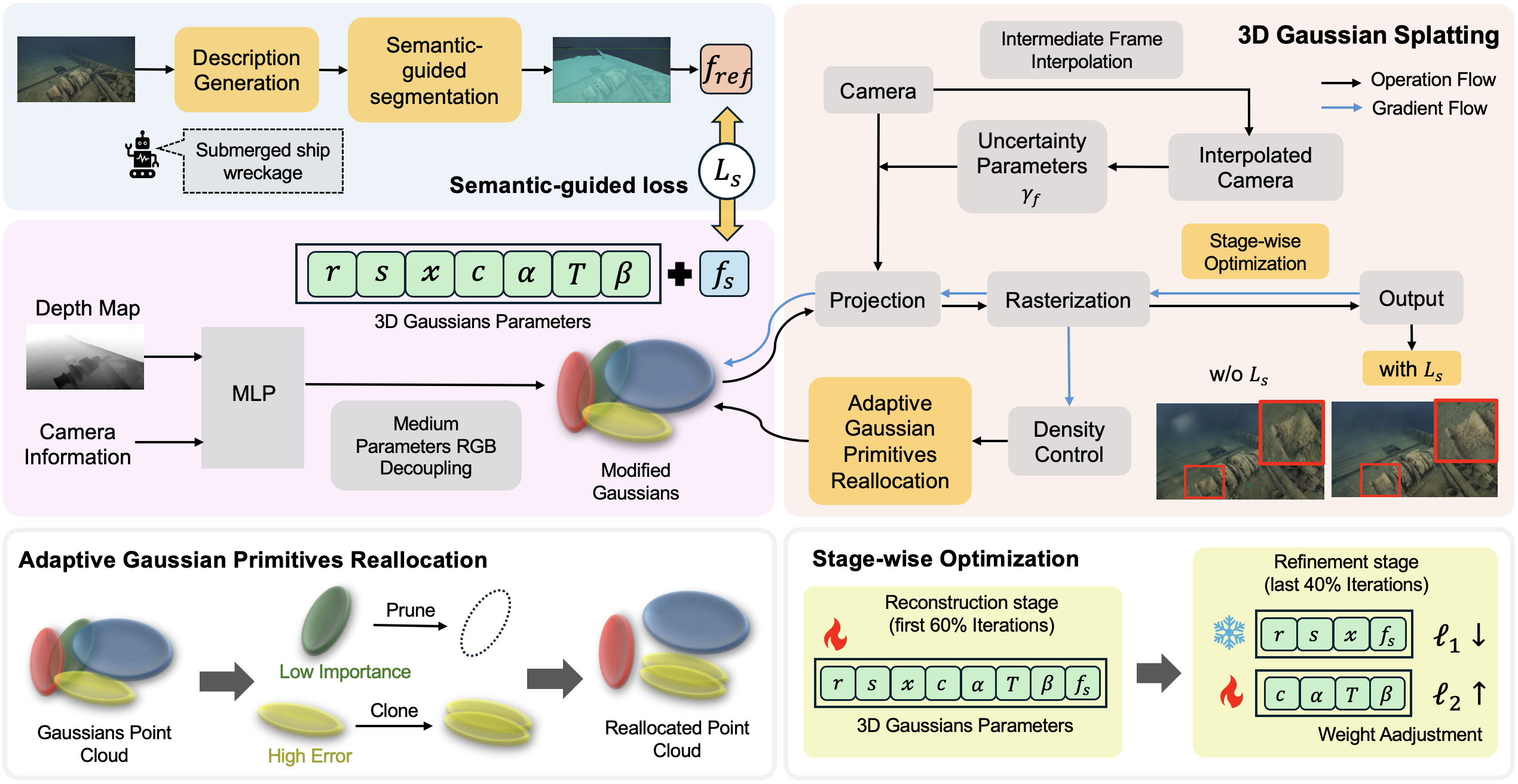}
  \caption{Pipeline of the proposed framework. Yellow highlights indicate key components: (1) a semantic-guided loss $L_s$ for object-aware regularization; (2) adaptive Gaussian primitives reallocation to improve representation balance under a fixed primitive budget; and (3) a stage-wise optimization for stable training and improved reconstruction fidelity.}
  \label{fig:overview}
\end{figure*}

% ---------------------
\section{Methodology}
\label{sec:method}

The overall framework is illustrated in \autoref{fig:overview}, where the new components are highlighted in yellow. Each Gaussian primitive is extended with a learnable semantic feature vector, which is aligned with embeddings derived from CLIP to incorporate high-level contextual information. A semantic consistency loss $L_s$, defined over region-level segmentation, provides auxiliary supervision that promotes object-level coherence and reduces the influence of redundant background regions. To address representation imbalance, an adaptive Gaussian allocation mechanism is introduced to redistribute primitives from low-contribution regions to areas with high reconstruction error. This process improves spatial coverage and reconstruction fidelity while maintaining a fixed primitive budget. In addition, a stage-wise optimization strategy is employed to improve training stability. The optimization proceeds from an initial stage focused on global structure to a refinement stage that emphasizes fine-grained appearance, enabling more consistent reconstruction under degraded imaging conditions.
The framework also incorporates the frame interpolation strategy from our previous work~\citep{jiang2025rusplatting}, which has been shown to improve performance in sparse-view scenarios.

% --------------------------------------
\subsection{Semantic-guided Gaussians}

Traditional 3D Gaussian Splatting optimizes all primitives uniformly using photometric reconstruction loss, without considering spatial variability in information content. This assumption is suboptimal in underwater environments, where visual quality is highly non-uniform due to scattering, attenuation, and sparse observations. Consequently, salient foreground structures require higher reconstruction fidelity, while degraded regions provide weak and unreliable signals, hindering geometric consistency.

\begin{figure*}[pos=bt]
    \centering
\includegraphics[width=\linewidth]{ 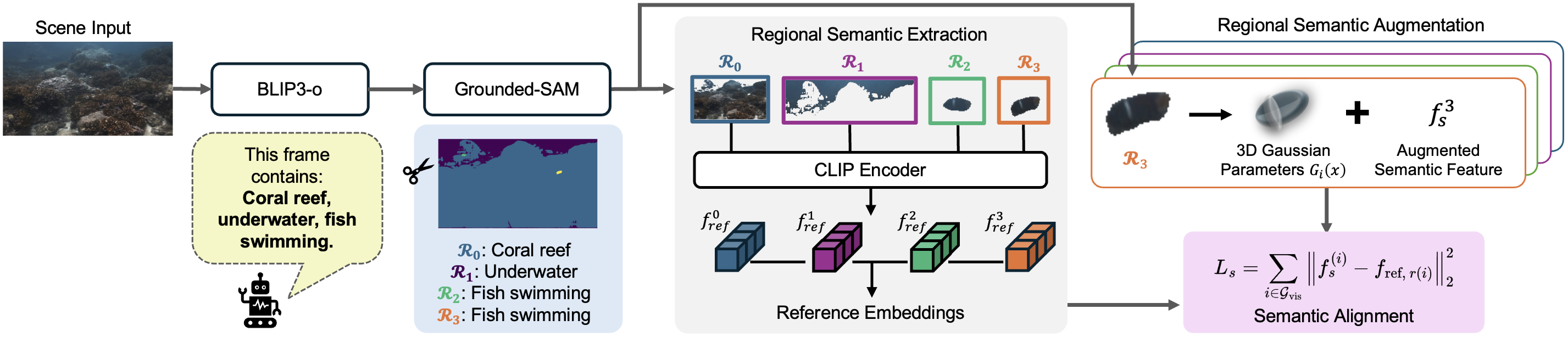}
    \caption{Overview of the proposed semantic-guided Gaussian module. BLIP3-o and Grounded-SAM are used to extract semantic regions, from which CLIP-based features are derived to supervise the learnable semantic features of the corresponding Gaussian primitives. }
    \label{fig:SGprocess}
\end{figure*}

To address this limitation, we incorporate semantic information into the Gaussian representation to enable object-aware regularization. An overview of the proposed semantic-guided Gaussian module is shown in \autoref{fig:SGprocess}. By integrating semantic features, the framework links high-level scene understanding with low-level degradation. In turbid conditions, photometric supervision alone struggles to distinguish true geometry from medium-induced artifacts such as backscatter. The proposed semantic-guided loss ($L_s$) acts as a structural constraint by enforcing feature consistency among Gaussian primitives within the same semantic region (e.g., coral reefs or shipwrecks). This discourages the allocation of density to semantically incoherent regions such as floating particles or haze, thereby improving the fidelity of reconstructed geometry, particularly around salient object boundaries.

In the proposed framework, each Gaussian primitive is extended with an additional semantic attribute represented by a learnable feature vector $f_s \in \mathbb{R}^d$, where $d$ denotes the dimensionality of the semantic embedding space. In contrast to geometric and photometric attributes, $f_s$ is optimized under external supervision to encode high-level contextual information, enabling object-aware regularization during reconstruction.

To obtain reference semantic embeddings, we first generate textual descriptions for each scene using BLIP3-o~\citep{chen2025blip3}. These descriptions are then used to guide Grounded-SAM~\citep{ren2024grounded} to identify regions of interest in the input image $I_{\mathrm{img}}$. The process can be expressed as:
\begin{equation}
\begin{aligned}
R &= \{\mathcal{R}_k\}_{k=1}^{K} \\
           &= \text{Grounded-SAM}(I_{\mathrm{img}}, \text{BLIP3-o}(I_{\mathrm{img}})), \\
I_{\mathcal{R}_k} &= I_{\mathrm{img}}[\mathcal{R}_k], \\
f_{\text{ref},k} &= \text{CLIP}(I_{\mathcal{R}_k}), \qquad k=1,\dots,K.
\end{aligned}
\end{equation}
where $R = \{\mathcal{R}_k\}_{k=1}^{K}$ denotes the set of detected semantic regions, $\mathcal{R}_k$ denotes the $k$-th semantic region, $I_{\mathcal{R}_k}$ represents the cropped image patch corresponding to the $k$-th region, and $f_{\text{ref},k}$ denotes the CLIP embedding extracted from that region. In practice, a fixed number of regions ($K=3$ in our implementation) is selected for each image to cover the dominant semantic content in the scene. Each selected region provides a region-level semantic target for supervising the Gaussian primitives whose projections fall within that region.

During training, Gaussian primitives whose projections fall within the detected semantic regions are encouraged to align their semantic features with the corresponding reference embedding via the following loss:
\begin{equation}
L_s = \sum_{i \in \mathcal{G}_{\mathrm{vis}}} \left\| f_s^{(i)} - f_{\text{ref},\,r(i)} \right\|_2^2,
\label{semantic_loss}
\end{equation}
where $\mathcal{G}_{\mathrm{vis}}$ denotes the set of visible Gaussian primitives assigned to semantic regions, and $r(i)$ denotes the index of the semantic region associated with the projection of the $i$-th Gaussian. Accordingly, $f_{\text{ref},\,r(i)}$ is the reference semantic embedding assigned to the $i$-th Gaussian based on its corresponding region, and $f_s^{(i)}$ denotes the learnable semantic feature of the $i$-th Gaussian primitive $G_i$. In cases of overlapping regions, a deterministic first-match rule is applied, while Gaussians not covered by any region are excluded from semantic supervision. The loss is applied in a view-dependent manner, such that visibility and occlusion are handled implicitly by the standard 3DGS rendering pipeline.

The semantic term is designed as an auxiliary regularization rather than a primary supervision signal. To this end, region assignments and target embeddings are treated as fixed during optimization, and semantic features are not directly involved in the rendering process. This design mitigates the impact of noisy or imperfect semantic cues, which are common in underwater imagery, and ensures that geometric optimization remains driven primarily by photometric consistency while benefiting from additional semantic guidance.

This additional supervision promotes semantic–geometric consistency by encouraging Gaussian primitives within the same region to exhibit similar semantic features, thereby supporting object-level coherence under noisy, low-visibility, or sparse-view conditions, and improving reconstruction consistency in regions where photometric cues are degraded. As a result, the reconstructed scenes demonstrate improved structural consistency and perceptual quality.

% ---------------------Haoran--------------------------------------------
\begin{figure*}[pos=bt]
  \centering
  \includegraphics[width=\textwidth]{ 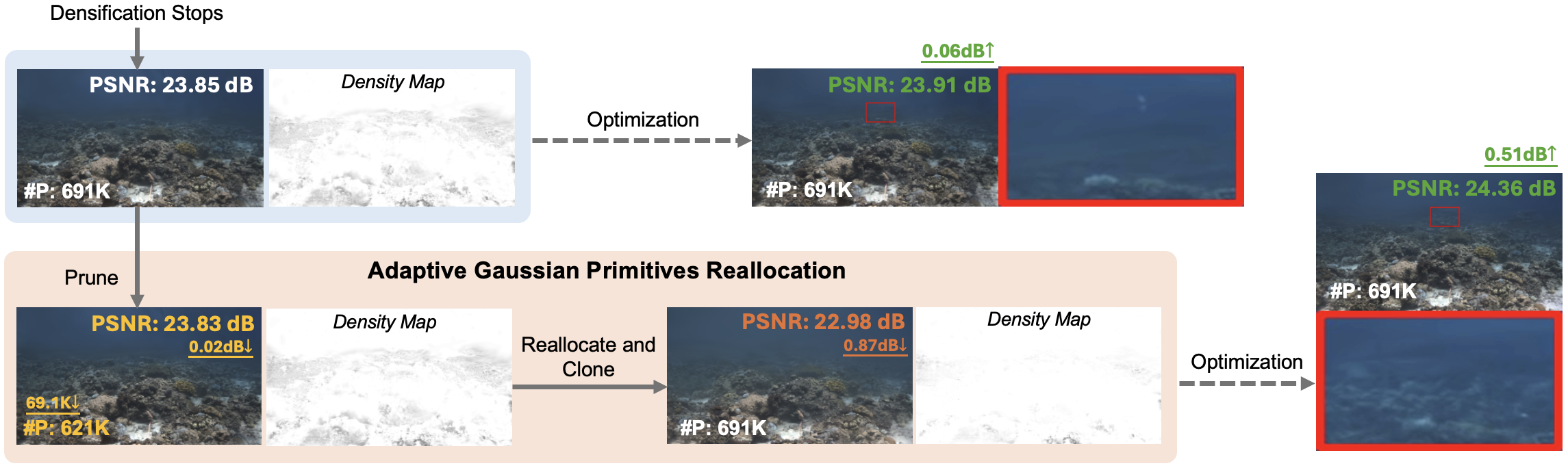}
  \caption{Comparison of reconstruction results without (top) and with (bottom) the adaptive Gaussian primitive reallocation process. The reallocation produces a more balanced spatial distribution of primitives, as illustrated by the corresponding normalized density maps, and leads to improved reconstruction quality, reflected by higher PSNR values after optimization. In the normalized density maps, darker regions indicate lower Gaussian density.}
  \label{fig:Reallocation}
\end{figure*}

\subsection{Adaptive Gaussian Primitives Reallocation}

Existing 3DGS-based approaches employ gradient-driven densification, where position gradients are used to introduce additional Gaussian primitives during optimization. While effective for improving local detail, this strategy can lead to redundancy in the representation~\citep{fang2024mini} and may limit overall reconstruction quality.
To address this limitation, we introduce an adaptive Gaussian primitive reallocation strategy that redistributes primitives from low-contribution regions to areas with high reconstruction error, thereby improving representation efficiency and spatial coverage. 

Similar to the Speedy-Splat~\citep{hanson2025speedy}, we define an importance score $\tilde{S}_i$ for each Gaussian primitive $i$ based on its contribution to the rendered image. Specifically,
\begin{equation}
    \tilde{S}_i = \log \left| \nabla_{I_G} g_i \, \nabla_{I_G} g_i^{T} \right|,
    \label{eq:prune_score_log}
\end{equation}
where $I_G$ denotes the rendered image formed by all Gaussians, and $\nabla_{I_G} g_i$ represents the gradient of the contribution of the $i$-th Gaussian with respect to $I_G$.  

Since $g_i$ is scalar-valued and the logarithm is a monotonically increasing function, the importance score can be simplified to
\begin{equation}
    \tilde{S}_i = \left( \nabla_{I_G} g_i \right)^2,
    \label{eq:prune_score_sq}
\end{equation}
which provides a measure of the sensitivity of the rendered image to each Gaussian primitive. Higher values of $\tilde{S}_i$ indicate that the corresponding primitive has a stronger influence on the rendered image.

To further improve reconstruction quality, we define an error score $\tilde{E}_i$ that quantifies the contribution of each Gaussian primitive to the reconstruction loss $L_{\text{Rec}}$:
\begin{equation}
    \tilde{E}_i = \left( \nabla_{L_{\text{Rec}}} g_i \right)^2,
    \label{eq:prune_score_sq_2}
\end{equation}
where $\nabla_{L_{\text{Rec}}} g_i$ denotes the gradient of the $i$-th Gaussian contribution with respect to the reconstruction loss.

After the densification stage, the importance score $\tilde{S}_i$ and error score $\tilde{E}_i$ are evaluated periodically (every 3,000 iterations in this work). Based on $\tilde{S}_i$, a fraction of primitives with the lowest scores (bottom 10\%) is removed, as these contribute minimally to the rendered output. The corresponding budget is then reallocated to primitives with the highest error scores (top 10\% ranked by $\tilde{E}_i$) through a cloning operation. Specifically, new primitives are initialized with identical attributes to the selected high-error Gaussians and are subsequently refined through optimization to better capture local structure.
The pruning and cloning ratio is defined heuristically; in practice, stable performance is observed within a reasonable range, and consistent improvements are demonstrated in the ablation study (\autoref{tab:ablation}).  

\autoref{fig:Reallocation} illustrates the effect of the adaptive Gaussian reallocation process (bottom branch) compared to standard densification (top branch). Every 3,000 iterations, after densification has stopped, reallocation begins with a pruning step that removes 10\% of low-importance primitives. This pruning stage leads to a slight reduction in reconstruction quality (approximately 0.02 dB in the example) due to the temporary decrease in representation capacity. Subsequently, the reallocation process redistributes the freed primitives to regions with higher reconstruction error, resulting in a more balanced spatial distribution. This effect is evident in the corresponding normalized density maps, derived from the projected opacity $\alpha_i$, which exhibit more uniform coverage after reallocation. Following further optimization, the reconstruction quality improves significantly compared to the case without reallocation, as shown by the two results on the right of \autoref{fig:Reallocation}, where previously high-error regions exhibit clearer structural details after Gaussian reallocation.

Overall, our adaptive reallocation strategy acts as a dynamic density management mechanism that improves representation efficiency by transferring capacity from low-contribution regions to areas with high reconstruction error. This alleviates the imbalance introduced by gradient-driven densification, reduces redundancy, and mitigates overfitting, while maintaining a fixed number of Gaussian primitives throughout optimization, preserving computational efficiency for real-time rendering.

% --------------------------------------

\subsection{Stage-wise Optimization Strategy}

To improve optimization stability and reconstruction quality, we employ a two-stage coarse-to-fine training strategy (\autoref{fig:overview}, bottom-right panel). 
\begin{itemize}[leftmargin=*]
    \item \textbf{Stage 1} (Reconstruction stage): Focuses on jointly optimizing global geometry and appearance. The loss for Stage 1 is dominated by an $\ell_1$-based reconstruction term with SSIM, which is robust to noise. Additional loss terms are incorporated to preserve structural consistency and model underwater imaging physics, including depth alignment, medium-related regularization, edge smoothness, semantic alignment, and a small $\ell_2$ term. This stage promotes robust recovery of the global scene structure under degraded observations.
    \item \textbf{Stage 2} (Refinement stage): Freezes the geometric and semantic Gaussian parameters, including position $f_{xyz}$, rotation $f_{\text{rotation}}$, scale $f_{\text{scale}}$ and semantic feature $f_s$, focusing on appearance refinement. Specifically, appearance-related parameters continue to be optimized, while the semantic, depth, and smoothness regularization terms are disabled, the SSIM term is removed, and the $\ell_2$ reconstruction term is emphasized at this stage to enhance fine details and color consistency.
\end{itemize}

This stage-wise optimization scheme improves convergence under noisy and sparse-view conditions, reduces sensitivity to medium-induced degradation, and leads to more consistent reconstruction quality across challenging underwater scenes. In experiments, we define the reconstruction stage as the first 60\% of the iterations and the refinement stage as the final 40\% of the total iterations. This split serves as a practical heuristic, and the optimal transition point may vary across datasets and scene characteristics.

% --------------------------------------
\subsection{Loss Functions}

The training process follows a two-stage optimization schedule described above and guided by a composite objective function. The baseline losses, including the reconstruction loss $L_{\text{Rec}}$, depth supervision $L_{\text{Depth}}$, gray-world prior $L_g$, and edge-aware smoothness $L_{\text{Smooth}}$, are adopted from our previous work~\citep{jiang2025rusplatting}.
To improve stability and appearance modeling, additional loss terms are introduced in this paper. These include the semantic loss $L_s$ (Eq.~\ref{semantic_loss}), a pixel-wise mean squared error term $L_2$ to refine appearance, and a hinge loss $L_h$ that constrains the predicted attenuation coefficients $\beta^d$ and $\beta^b$. 

The overall objective function is formulated as
\begin{equation}
L_{\text{final}} = L_{\text{Rec}} + L_{\text{Depth}} + L_{g} + L_{\text{Smooth}} + \lambda_s L_s + \lambda_2 L_2 + \lambda_h L_h,
\end{equation}
where $\lambda_s$, $\lambda_2$, and $\lambda_h$ are weighting parameters controlling the contributions of the semantic, mean square error, and hinge terms, respectively. For interpolated frames, we adopt an uncertainty-weighted formulation:
\begin{equation}
L_{\text{final}}' = \tfrac{1}{2} \gamma L_{\text{final}} - \tfrac{1}{2} \zeta \log(\gamma),
\end{equation}
where $\gamma$ denotes the learned uncertainty and $\zeta$ is a regularization coefficient, as detailed in our previous work~\citep{jiang2025rusplatting}.

\subsubsection{Base losses}

$L_{\text{Rec}}$: The reconstruction loss combines a mean absolute error term, which is robust to outliers, with a structural similarity (SSIM) term:
\begin{equation}
    L_{\text{Rec}} = \lambda_1 (1 - \lambda_r) \left\| I_{\text{out}} - I_{\text{gt}} \right\|_1 + \lambda_r \bigl(1 - \text{SSIM}(I_{\text{out}}, I_{\text{gt}})\bigr),
\end{equation}
where $\lambda_1$ is a dynamic weight defined as $\lambda_1 = 0.7 + 0.5 \cdot (\text{iteration}/\text{iterations})$, and $\lambda_r$ controls the contribution of the SSIM term. Specifically, $\lambda_r$ is set to 0.2 in Stage 1 and 0 in Stage 2.

$L_{\text{Depth}}$: Depth $z$ plays an important role, as $z$ is used to estimate underwater medium parameters. We therefore introduce a depth supervision term:
\begin{equation}
    L_{\text{Depth}} = \lambda_d \left\| \hat{D} - D \right\|_1 + \lambda_{ca} \sum_{ch \in \{r,g,b\}} \sum_{n \in \{d,b\}} \left\| z^{n}_{ch} - D \right\|_1,
\end{equation}
where $\lambda_d$ and $\lambda_{ca}$ are weighting parameters, set to 0.1 and 1 in Stage 1, respectively, and both to 0 in Stage 2. Here, $D$ denotes the reference depth, and $\hat{D}$ represents the rendered depth. Since ground-truth depth is unavailable, pseudo-depth estimated using Depth Anything V2~\citep{yang2024depthv2} is used for supervision. Notably, underwater scenes were included as one of the scenarios in its benchmark. The term $z^{n}_{ch}$ represents the inferred depth under the direct ($n = d$) and backscatter ($n = b$) components for each color channel $ch$.

$L_g$: The gray-world prior is applied to the estimated scene radiance before underwater degradation, enforcing approximately equal mean intensities across RGB channels. The target value of $0.5$ corresponds to an idealized balanced intensity under the gray-world assumption. It serves to prevent the model from converging to implausible color casts during restoration. The loss is defined as
\begin{equation}
    L_g = \sum_{ch \in \{r, g, b\}} \left( \mu(J_{ch}) - 0.5 \right)^2,
\end{equation}
where $J_{ch}$ denotes the restored pixel values of channel $ch$, normalized to the range $[0,1]$, and $\mu(\cdot)$ represents the mean intensity of each channel. 

$L_{\text{Smooth}}$: Penalizes depth discontinuities that do not align with image edges, i.e., it allows sharp depth edges at color edges but penalizes floating or noisy depth in smooth regions. Our edge-aware smoothness loss is defined as,
\begin{equation}
\begin{aligned}
L_{\text{Smooth}} = \lambda_m \sum_{i,j} \Big[
& \exp\!\bigl(-\lambda_b \left| \Delta_x D_{i,j} \right|\bigr)
  \left| \Delta_x I_{i,j} \right| \\
& {}+ \exp\!\bigl(-\lambda_b \left| \Delta_y D_{i,j} \right|\bigr)
  \left| \Delta_y I_{i,j} \right|
\Big],
\end{aligned}
\label{smooth_loss_eq}
\end{equation}
where $(i,j)$ indexes the spatial coordinates in the image plane, $I$ denotes the input image and $D$ denotes the pseudo depth map, $\Delta_x D_{i,j}=D_{i,j+1}-D_{i,j}$ and $\Delta_y D_{i,j}=D_{i+1,j}-D_{i,j}$ denote the forward differences of the depth map along the horizontal and vertical directions, respectively. Likewise, $\Delta_x I_{i,j}=I_{i,j+1}-I_{i,j}$ and $\Delta_y I_{i,j}=I_{i+1,j}-I_{i,j}$ are the corresponding image gradients. Here, $\lambda_b$ is fixed to 5, and $\lambda_m$ is set to 0.05 in Stage 1 and 0 in Stage 2.

\begin{figure}
\includegraphics[width=\linewidth]{ 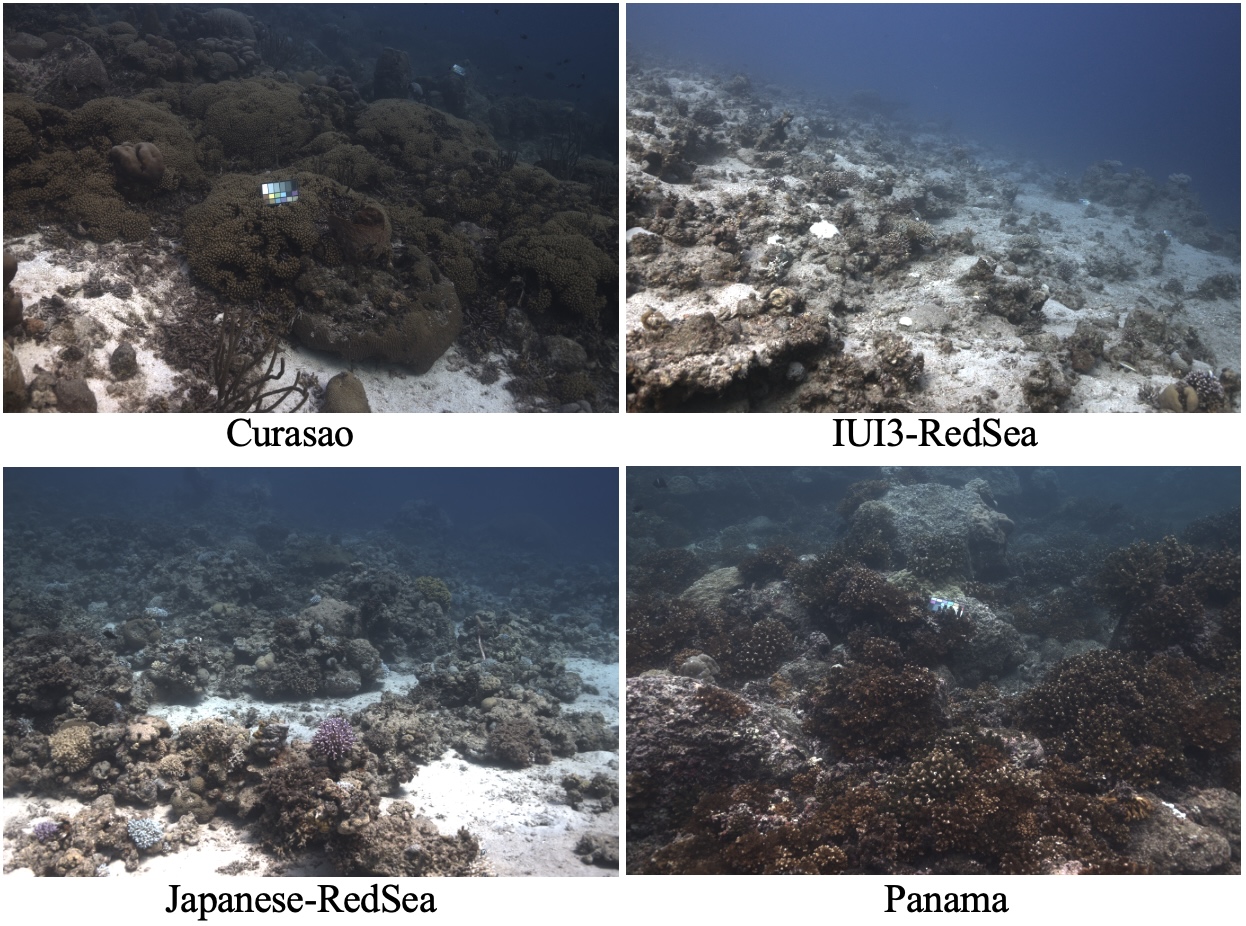} \vspace{-2mm}
  \caption{Sample images of each scene from \textbf{SeaThru-NeRF} dataset~\citep{levy2023seathru}.} \vspace{1pt}
  \label{fig:seathreNerf} 
%\end{figure}
%\begin{figure}
  \includegraphics[width=\linewidth]{ 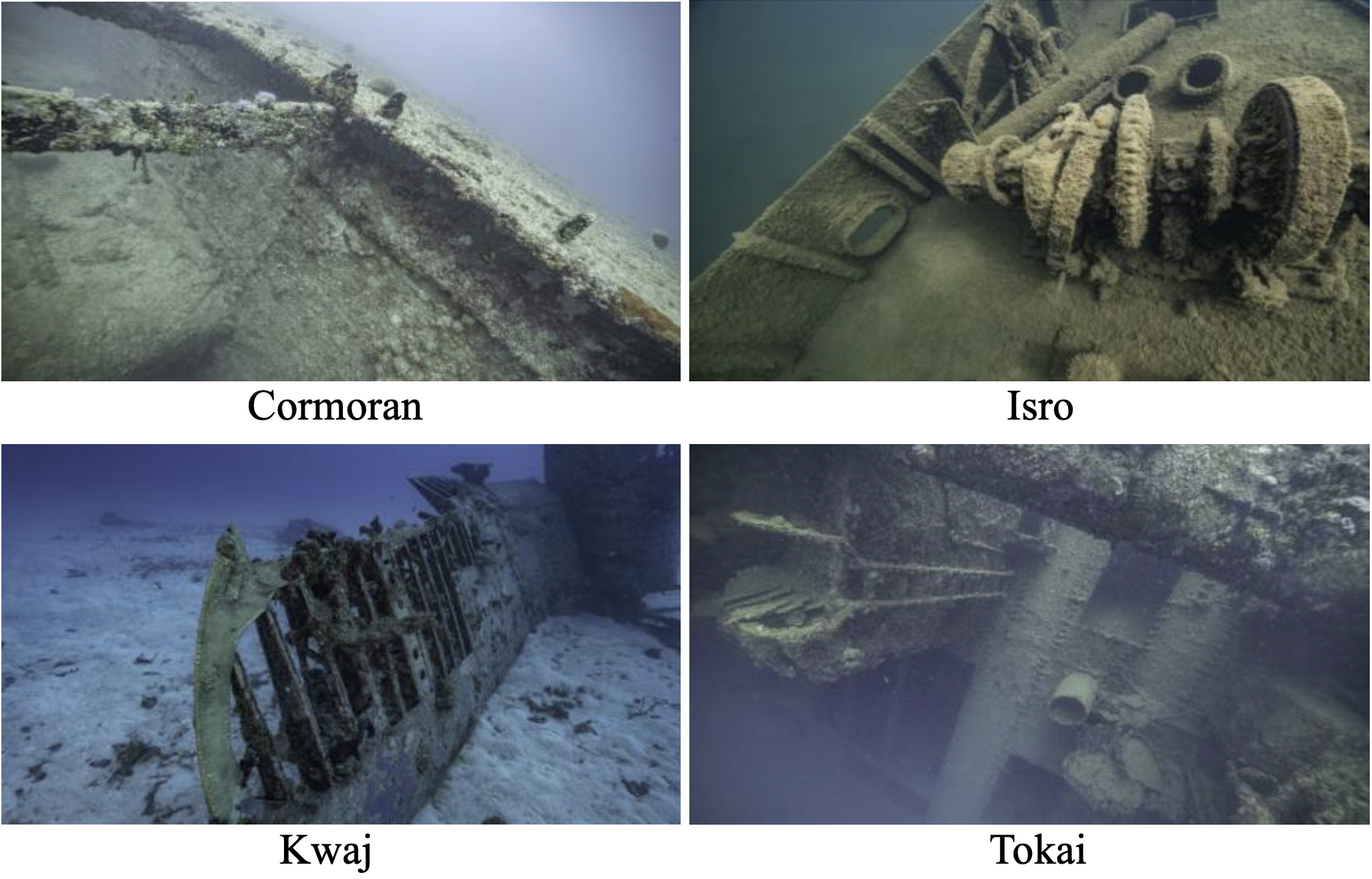} \vspace{-2mm}
  \caption{Sample images of each scene from \textbf{Submerged3D} dataset ~\citep{jiang2025rusplatting}.}  \vspace{1pt}
  \label{fig:submerge3D}
%\end{figure}
%\begin{figure}
  \includegraphics[width=\linewidth]{ 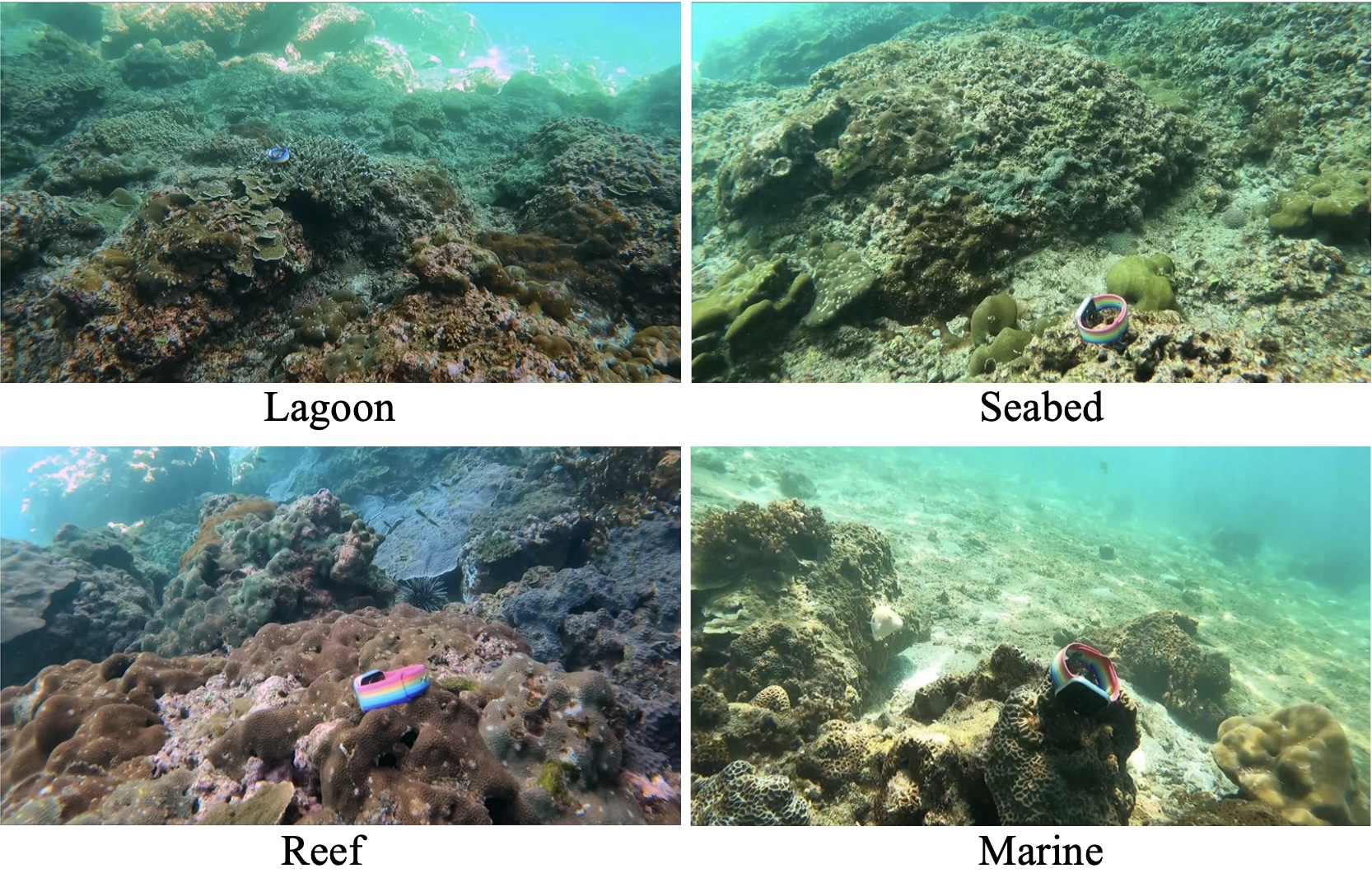} \vspace{-2mm}
  \caption{Sample images of each scene from \textbf{S-UW} dataset ~\citep{wang2025uw}.}
  \label{fig:SUW}
\end{figure}

\subsubsection{Additional losses}

This section describes three new loss terms introduced in this paper.

$L_s$: The semantic loss is defined in Eq.~\ref{semantic_loss}. Its weighting coefficient $\lambda_s$ is scheduled during training in Stage 1 as
\[
\lambda_s = \max\left(0.01,\; 0.3 \cdot \left(1 - \frac{\text{iteration}}{\text{iterations}}\right)\right).
\]
This assigns a higher weight in the early stages of optimization, when global structure is being established, and gradually reduces its influence as training progresses. In Stage 2, the semantic loss is disabled ($\lambda_s = 0$) to allow fine-grained appearance refinement without additional semantic constraints.

$L_2$: The mean squared error term is introduced to enhance sensitivity to high-frequency details and high-contrast regions. During Stage 1, it is assigned a low weight, increasing gradually according to $0.1 \cdot \text{iteration}/\text{iterations}$. In Stage 2, its contribution is increased to 0.9 to support fine-grained appearance refinement. This transition from $\ell_1$-dominated to $\ell_2$-dominated optimization reflects the shift from global structure learning to precise color and detail reconstruction once the geometry has stabilized.

$L_h$: The hinge term acts as a weak regularization ($\lambda_h=0.001$) that encourages a consistent ordering of attenuation across color channels, reflecting typical underwater imaging behavior while avoiding strict physical constraints. As the coefficients are estimated per scene, only their relative ordering is enforced, allowing flexibility under varying environmental conditions. 

For a desired inequality $a>b$, we define a hinge loss as
\begin{equation}
L_{\text{hinge}}(a,b) = \max(0,\, b - a + m),
\end{equation}
where $m \ge 0$ is a small margin (set to $10^{-3}$). The loss is zero when the constraint is satisfied and increases linearly when violated.  

Following the wavelength-dependent attenuation behavior of underwater light propagation, where longer wavelengths attenuate more rapidly than shorter ones, we impose an ordering constraint on the attenuation coefficients such that $\beta^{*}_r > \beta^{*}_g > \beta^{*}_b$ for $* \in \{d, b\}$. This is implemented as
\begin{equation}
L_h^{*} = \max(0, \beta^{*}_g - \beta^{*}_r + m) + \max(0, \beta^{*}_b - \beta^{*}_g + m),
\end{equation}
where subscripts denote the RGB channels. While this ordering reflects typical attenuation behavior in clear water, deviations may occur in certain environments (e.g., chlorophyll-rich coastal waters), and this can be straightforwardly adjusted. The hinge term therefore acts as a weak regularization rather than a strict constraint.

The final hinge loss is defined as
\begin{equation}
L_h = \tfrac{1}{2}(L_h^{d} + L_h^{b}).
\end{equation}

In practice, we found that integrating $L_h$ into the optimization can improve the rendering quality by approximately 0.066 dB. However, it also improves training stability, as observed from the total training loss and, in particular, $L_{Rec}$.

% ----------------------------------------------------
\begin{table*}[!t]
    \caption{Performance comparison between SWAGSplatting and six baseline models (Instant-NGP~\citep{muller2022instant}, SeaThru-NeRF~\citep{levy2023seathru}, 3DGS~\citep{kerbl:3Dgaussians:2023}, WaterSplatting~\citep{li2024watersplatting}, UW-GS~\citep{wang2025uw}, and RUSplatting~\citep{jiang2025rusplatting} ) on three datasets (SeaThru-NeRF, S-UW and Submerged3D). $\uparrow$ indicates that higher values are better, while $\downarrow$ indicates that lower values are better.\setlength{\fboxsep}{1pt} \colorbox{lightred}{\strut Red}, \colorbox{lightorange}{\strut orange}, and \colorbox{lightyellow}{\strut yellow} denote the best, second-best, and third-best results, respectively.}
    %==================== SeaThru-NeRF ====================
    \resizebox{\textwidth}{!}{
    \begin{tabular}{r|ccc|ccc|ccc|ccc|ccc}
        \toprule
        \multicolumn{16}{c}{\textbf{SeaThru-NeRF}} \\
        \midrule
        Scene &
        \multicolumn{3}{c|}{Curacao} &
        \multicolumn{3}{c|}{Panama} &
        \multicolumn{3}{c|}{IUI-Redsea} &
        \multicolumn{3}{c|}{Japanese-Redsea} &
        \multicolumn{3}{c}{Average} \\
        \midrule
        Method &
        PSNR$\uparrow$ & SSIM$\uparrow$ & LPIPS$\downarrow$ &
        PSNR$\uparrow$ & SSIM$\uparrow$ & LPIPS$\downarrow$ &
        PSNR$\uparrow$ & SSIM$\uparrow$ & LPIPS$\downarrow$ &
        PSNR$\uparrow$ & SSIM$\uparrow$ & LPIPS$\downarrow$ &
        PSNR$\uparrow$ & SSIM$\uparrow$ & LPIPS$\downarrow$ \\
        \midrule
        Instant-NGP &
        27.9215 & 0.7013 & 0.3961 &
        23.4341 & 0.6294 & 0.4361 &
        20.6162 & 0.5594 & 0.6121 &
        23.2061 & 0.6371 & 0.3294 &
        23.7945 & 0.6318 & 0.4434 \\

        SeaThru-NeRF &
        29.9162 & 0.8463 & 0.2984 &
        26.9634 & 0.7261 & 0.3684 &
        25.4989 & 0.7934 & 0.3684 &
        21.8631 & 0.7569 & 0.3243 &
        26.0604 & 0.7807 & 0.3399 \\

        3DGS &
        29.0234 & 0.9127 & 0.1790 &
        30.7038 & 0.7606 & 0.1560 &
        25.2240 & 0.9005 & 0.1970 &
        22.3902 & 0.8638 & 0.1960 &
        26.8353 & 0.8594 & 0.1820 \\

        WaterSplatting &
        \cellcolor{lightorange}31.4616 & \cellcolor{lightyellow}0.9256 & \cellcolor{lightorange}0.1560 &
        \cellcolor{lightyellow}31.5367 & 0.9213 & 0.1445 &
        22.1392 & 0.6598 & 0.2984 &
        \cellcolor{lightorange}24.4323 & \cellcolor{lightorange}0.8834 & \cellcolor{lightred}0.1494 &
        27.3925 & 0.8475 & 0.1871 \\

        UW-GS &
        29.9721 & 0.9220 & 0.1770 &
        31.0023 & \cellcolor{lightyellow}0.9335 & \cellcolor{lightyellow}0.1432 &
        \cellcolor{lightyellow}28.5473 & \cellcolor{lightyellow}0.9243 & \cellcolor{lightyellow}0.1885 &
        23.5013 & 0.8625 & 0.1916 &
        \cellcolor{lightyellow}28.2558 & \cellcolor{lightyellow}0.9106 & \cellcolor{lightyellow}0.1751 \\

        RUSplatting &
        \cellcolor{lightyellow}30.9557 & \cellcolor{lightorange}0.9318 & \cellcolor{lightyellow}0.1611 &
        \cellcolor{lightorange}31.8683 & \cellcolor{lightorange}0.9339 & \cellcolor{lightorange}0.1396 &
        \cellcolor{lightorange}29.8036 & \cellcolor{lightorange}0.9292 & \cellcolor{lightorange}0.1847 &
        \cellcolor{lightred}24.5436 & \cellcolor{lightyellow}0.8706 & \cellcolor{lightyellow}0.1809 &
        \cellcolor{lightorange}29.2928 & \cellcolor{lightorange}0.9164 & \cellcolor{lightorange}0.1666 \\

        \midrule
        Ours &
        \cellcolor{lightred}33.0521 & \cellcolor{lightred} 0.9481 & \cellcolor{lightred}0.1428 &
        \cellcolor{lightred}32.0630 & \cellcolor{lightred} 0.9389 & \cellcolor{lightred}0.1264 &
        \cellcolor{lightred}30.3602 & \cellcolor{lightred}0.9306 & \cellcolor{lightred}0.1822 &
        \cellcolor{lightyellow}24.3617 & \cellcolor{lightred}0.8861 & \cellcolor{lightorange}0.1772 &
        \cellcolor{lightred}\textbf{29.9593} & \cellcolor{lightred}\textbf{0.9260} & \cellcolor{lightred}\textbf{0.1571} \\
        % \midrule
    % \end{tabular}
    % }
    % \vspace{1.5ex}
    % %==================== Submerged3D ====================
    % \resizebox{\textwidth}{!}{
    % \begin{tabular}{l|ccc|ccc|ccc|ccc|ccc|}
        \midrule
         \multicolumn{16}{c}{\textbf{Submerged3D}} \\
        \midrule
        Scene &
        \multicolumn{3}{c|}{Cormoran} &
        \multicolumn{3}{c|}{Isro} &
        \multicolumn{3}{c|}{Kwaj} &
        \multicolumn{3}{c|}{Tokai} &
        \multicolumn{3}{c}{Average} \\
        \midrule
        Method &
        PSNR$\uparrow$ & SSIM$\uparrow$ & LPIPS$\downarrow$ &
        PSNR$\uparrow$ & SSIM$\uparrow$ & LPIPS$\downarrow$ &
        PSNR$\uparrow$ & SSIM$\uparrow$ & LPIPS$\downarrow$ &
        PSNR$\uparrow$ & SSIM$\uparrow$ & LPIPS$\downarrow$ &
        PSNR$\uparrow$ & SSIM$\uparrow$ & LPIPS$\downarrow$ \\
        \midrule
        Instant-NGP &
        19.5271 & 0.5055 & 0.5505 &
        20.1861 & 0.6588 & 0.4146 &
        18.1444 & 0.4682 & 0.4535 &
        18.2488 & 0.4082 & 0.5723 &
        19.0266 & 0.5102 & 0.4977 \\

        SeaThru-NeRF &
        16.8816 & 0.4533 & 0.6118 &
        21.5887 & 0.6501 & 0.5025 &
        20.7094 & 0.5116 & 0.5240 &
        17.6072 & 0.3720 & 0.6360 &
        19.1967 & 0.4968 & 0.5686 \\

        3DGS &
        20.4698 & 0.6756 & 0.3852 &
        22.8198 & 0.8421 & 0.3008 &
        24.0964 & 0.8485 & 0.2276 &
        23.2345 & 0.6517 & 0.4017 &
        22.6551 & 0.7545 & 0.3288 \\

        WaterSplatting &
        \cellcolor{lightyellow}21.0749 & 0.6482 & 0.3846 &
        \cellcolor{lightyellow}26.2920 & 0.8517 & 0.2716 &
        27.8613 & 0.8709 & \cellcolor{lightyellow}0.2162 &
        23.1047 & 0.6362 & 0.3965 &
        \cellcolor{lightyellow}24.5832 & 0.7518 & 0.3172 \\

        UW-GS &
        20.6942 & \cellcolor{lightyellow}0.6800 & \cellcolor{lightred}0.3412 &
        24.6752 & \cellcolor{lightyellow}0.8649 & \cellcolor{lightyellow}0.2683 &
        \cellcolor{lightyellow}28.4065 & \cellcolor{lightred}0.8868 & \cellcolor{lightred}0.2105 &
        \cellcolor{lightyellow}23.5880 & \cellcolor{lightyellow}0.6578 & \cellcolor{lightyellow}0.3833 &
        24.3410 & \cellcolor{lightyellow}0.7724 & \cellcolor{lightorange}0.3008 \\

        RUSplatting &
        \cellcolor{lightorange}21.9625 & \cellcolor{lightorange}0.6846 & \cellcolor{lightyellow}0.3662 &
        \cellcolor{lightorange}27.8693 & \cellcolor{lightorange}0.8680 & \cellcolor{lightorange}0.2670 &
        \cellcolor{lightred}28.9058 & \cellcolor{lightyellow}0.8774 & 0.2253 &
        \cellcolor{lightorange}24.4585 & \cellcolor{lightorange}0.6599 & \cellcolor{lightred}0.3548 &
        \cellcolor{lightorange}25.7990 & \cellcolor{lightorange}0.7725 & \cellcolor{lightyellow}0.3033 \\
        \midrule
        Ours &
        \cellcolor{lightred}21.9794 & \cellcolor{lightred}0.7023 & \cellcolor{lightorange}0.3482 &
        \cellcolor{lightred}28.1544 & \cellcolor{lightred}0.8782 & \cellcolor{lightred}0.2643 &
        \cellcolor{lightorange}28.8320 & \cellcolor{lightorange}0.8847 & \cellcolor{lightorange}0.2161 &
        \cellcolor{lightred}25.6028 & \cellcolor{lightred}0.7391 & \cellcolor{lightorange}0.3612 &
        \cellcolor{lightred}\textbf{26.1422} & \cellcolor{lightred}\textbf{0.8011} & \cellcolor{lightred}\textbf{0.2974} \\
\midrule
        \multicolumn{16}{c}{\textbf{S-UW}} \\
        \midrule
        Scene &
        \multicolumn{3}{c|}{Lagoon} &
        \multicolumn{3}{c|}{Marine} &
        \multicolumn{3}{c|}{Reef} &
        \multicolumn{3}{c|}{Seabed} &
        \multicolumn{3}{c}{Average} \\
        \midrule
        Method &
        PSNR$\uparrow$ & SSIM$\uparrow$ & LPIPS$\downarrow$ &
        PSNR$\uparrow$ & SSIM$\uparrow$ & LPIPS$\downarrow$ &
        PSNR$\uparrow$ & SSIM$\uparrow$ & LPIPS$\downarrow$ &
        PSNR$\uparrow$ & SSIM$\uparrow$ & LPIPS$\downarrow$ &
        PSNR$\uparrow$ & SSIM$\uparrow$ & LPIPS$\downarrow$ \\
        \midrule
        
        Instant-NGP &
        21.6674 & 0.5480 & 0.3198 &
        15.7300 & 0.4637 & 0.4410 &
        18.5866 & 0.4382 & 0.5058 &
        19.6167 & 0.4275 & 0.3198 &
        18.9002 & 0.4694 & 0.3966 \\
        
        SeaThru-NeRF &
        25.5424 & 0.7592 & 0.2860 &
        18.1031 & 0.5718 & 0.4582 &
        18.5510 & 0.4077 & 0.5522 &
        22.7050 & 0.6407 & 0.3455 &
        21.2254 & 0.5949 & 0.4105 \\
        
        3DGS &
        27.1166 & \cellcolor{lightorange}0.8707 & \cellcolor{lightyellow}0.1606 &
        18.9492 & 0.6922 & 0.2990 &
        22.4663 & 0.7117 & 0.2646 &
        26.4933 & \cellcolor{lightyellow}0.8234 & 0.1684 &
        23.7564 & 0.7745 & 0.2232 \\
        
        WaterSplatting &
        \cellcolor{lightyellow}27.1979 & \cellcolor{lightyellow}0.8685 & \cellcolor{lightred}0.1135 &
        19.6173 & 0.6869 & \cellcolor{lightred}0.2619 &
        \cellcolor{lightyellow}26.6378 & \cellcolor{lightyellow}0.7971 & \cellcolor{lightred}0.1769 &
        26.0623 & 0.8136 & \cellcolor{lightred}0.1516 &
        \cellcolor{lightyellow}24.8788 & \cellcolor{lightyellow}0.7915 & \cellcolor{lightred}0.1760 \\
        
        UW-GS &
        26.9103 & \cellcolor{lightred}0.8736 & \cellcolor{lightorange}0.1479 &
        \cellcolor{lightyellow}20.0065 & \cellcolor{lightyellow}0.7145 & \cellcolor{lightorange}0.2678 &
        23.3078 & 0.7277 & 0.2504 &
        \cellcolor{lightyellow}26.5088 & 0.8197 & \cellcolor{lightorange}0.1541 &
        24.1834 & 0.7839 & \cellcolor{lightyellow}0.2051 \\
        
        RUSplatting &
        \cellcolor{lightred}27.6000 & 0.8659 & 0.1719 &
        \cellcolor{lightred}20.9329 & \cellcolor{lightred}0.7447 & \cellcolor{lightyellow}0.2706 &
        \cellcolor{lightorange}26.7588 & \cellcolor{lightred}0.8154 & \cellcolor{lightyellow}0.2222 &
        \cellcolor{lightorange}27.0903 & \cellcolor{lightorange}0.8292 & 0.1771 &
        \cellcolor{lightorange}25.5955 & \cellcolor{lightred}0.8138 & 0.2105 \\
        
        \midrule
        Ours &
        \cellcolor{lightorange}27.5294 & 0.8659 & 0.1625 &
        \cellcolor{lightorange}20.7229 & \cellcolor{lightorange}0.7328 & 0.2764 &
        \cellcolor{lightred}26.8547 & \cellcolor{lightorange}0.8127 & \cellcolor{lightorange}0.2124 &
        \cellcolor{lightred}27.4631 & \cellcolor{lightred}0.8370 & \cellcolor{lightyellow}0.1553 &
        \cellcolor{lightred}\textbf{25.6425} & \cellcolor{lightorange}\textbf{0.8121} & \cellcolor{lightorange}\textbf{0.2017} \\
        \bottomrule
    \end{tabular}
    }
    \label{tab:comparison}
\end{table*}
% ============================================================
% ============================================================

\section{Datasets}
We evaluate the proposed method on three public real-world underwater datasets, SeaThru-NeRF~\citep{levy2023seathru}, Submerged3D~\citep{jiang2025rusplatting} and S-UW~\citep{wang2025uw} . These datasets exhibit distinct characteristics of underwater imagery, providing complementary test conditions for assessing the robustness and generalization of 3D reconstruction methods. All images are resized to a resolution of 720p to ensure consistency across experiments. 

\textbf{SeaThru-NeRF} comprises four scenes captured across three marine environments, including the Red Sea (IUI3-RedSea and Japanese-RedSea), the Caribbean Sea (Curaçao), and the Pacific Ocean (Panama). The scenes contain 20, 20, and 18 RGB images, respectively, with three images per scene reserved for validation. The dataset covers a range of underwater imaging conditions and scene complexities. As shown in \autoref{fig:seathreNerf}, all scenes depict coral reef environments, with relatively limited variation in visibility and illumination.

\textbf{Submerged3D} comprises four scenes: the SMS Cormoran and Tokai Maru shipwrecks near Guam, the Kwajalein Atoll F4U-1 Corsair wreck near Mellu Island, and the SS America (Isro) near the Canary Islands. Each scene is derived from a larger dataset, from which 20 RGB images at 720p resolution are selected for processing. The data are provided by the National Park Service Submerged Resources Center\footnote{Cleaned photogrammetry models (with water removed) are available at \url{https://www.npssubmerged.com/photogrammetry}}.
The dataset presents significant challenges for 3D reconstruction due to sparse viewpoints, low-light conditions, and severe image degradation. Example scenes are shown in \autoref{fig:submerge3D}. For evaluation, we follow the predefined split, assigning every eighth image to the test set and using the remaining images for training.

\textbf{S-UW} consists of four scenes captured near the islands of Krabi, Thailand, with each scene comprising 24 RGB images. Acquired in shallow-water environments, the dataset exhibits additional challenges, including noticeable flickering caused by dynamic water surface motion (as illustrated in \autoref{fig:SUW}). The scenes predominantly feature coral reef structures with rich textures, while some sequences also include moving fish, introducing further complexity due to non-rigid motion.

\begin{figure*}[pos=bt]
  \includegraphics[width=\linewidth]{ 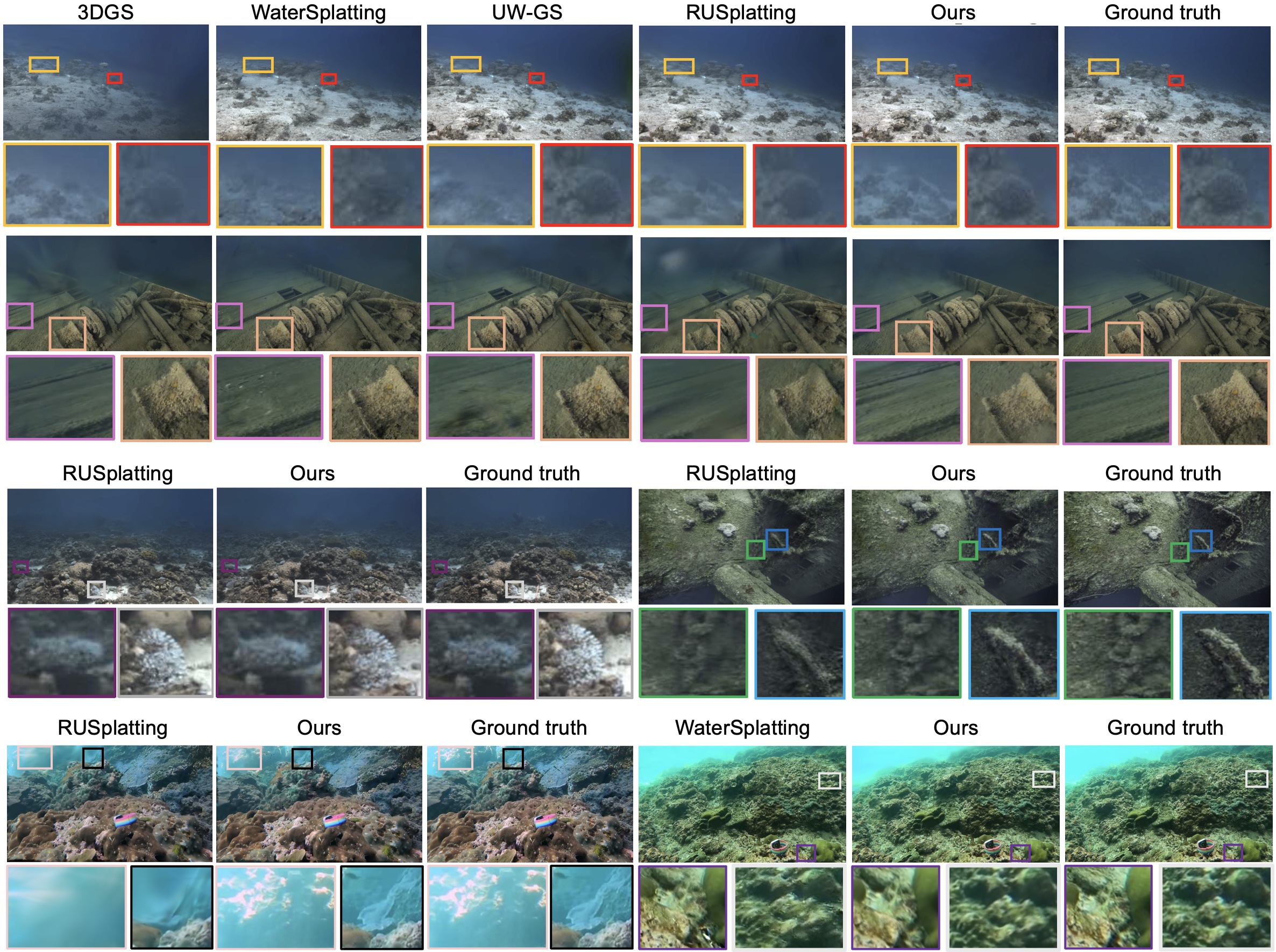}
  \caption{Novel view rendering comparison. The first row shows results from the IUI-Redsea scene from the SeaThru-NeRF dataset, and the second row shows the reconstructed scenes of the Isro from the Submerged3D dataset. The left side of the third row displays the reconstructed scene of the Japanese-Redsea, while the right side shows the Tokai from the Submerged3D. The fourth row is results from the S-UW dataset, showing the reconstructed scenes from Reef(left) and Seabed(right), respectively.}
  \label{overall_results}
\end{figure*}

\section{Experiment and Results}

\subsection{Experiment Setting}
All experiments are conducted on a single NVIDIA RTX 4090 GPU. We utilize COLMAP~\citep{schonberger2016structure} to initialize the point cloud and estimate camera poses for the image sequences. We modify the diff-gaussian-rasterization module used in the original 3DGS to render depth maps, incorporating an additional gradient computation branch in the backward pass for the depth loss $L_\text{Depth}$. The hyperparameter $\lambda_h$ is fixed to $1\times10^{-3}$, while $\lambda_s$ and $\lambda_2$ are dynamically scheduled during training. Specifically, $\lambda_s$ decays from $0.3$ to $0.01$ before being set to $0$ in the finetuning stage, whereas $\lambda_2$ increases linearly from $0$ to $0.1$ and is then set to $0.9$ during finetuning. To estimate 3DGS parameters, we employ an MLP with five layers, selected based on a hyperparameter sensitivity analysis.

Each model is trained for 20,000 iterations, with the second optimization stage commencing at iteration 12,000. To improve computational efficiency, the reference semantic embeddings $\{f_{\text{ref},k}\}_{k=1}^{K}$ are precomputed at the start of training and cached for reuse. The dimensionality of the projected CLIP embedding space is set to $d = 32$.
For frame interpolation, we adopt Real-Time Intermediate Flow Estimation (RIFE)~\citep{huang2022real}, following RUSplatting. A fixed weighting factor of 0.1 is used in place of adaptive weighting, providing a stable trade-off between reconstruction accuracy and interpolation quality across datasets.
In practice, we observed stable convergence across a range of values, and performance drops when either term is removed, indicating that both modules contribute.

We compare the proposed method against six representative baselines: Instant-NGP~\citep{muller2022instant}, SeaThru-NeRF~\citep{levy2023seathru}, 3DGS~\citep{kerbl:3Dgaussians:2023}, WaterSplatting~\citep{li2024watersplatting}, UW-GS~\citep{wang2025uw}, and RUSplatting~\citep{jiang2025rusplatting}. Several recent approaches, including Aquatic-GS~\citep{liu2024aquatic}, RecGS~\citep{zhang2025recgs}, R-Splatting~\citep{huang2025fromrestoration}, and AtlantisGS~\citep{Yi:AtlantisGS:2025}, are not publicly available and are therefore not included. Although SeaSplat~\citep{yang2025seasplat} provides official code, we observed numerical instability during training under the recommended settings, and it is thus excluded from comparison.

For fair evaluation, all baselines are implemented using their official codebases and trained on identical image sequences and optimization settings, following the same dataset split protocol. The loss weights are adopted from prior work to ensure a fair comparison.

\begin{figure*}[pos=bt]
    \centering
    \includegraphics[width=\linewidth]{ 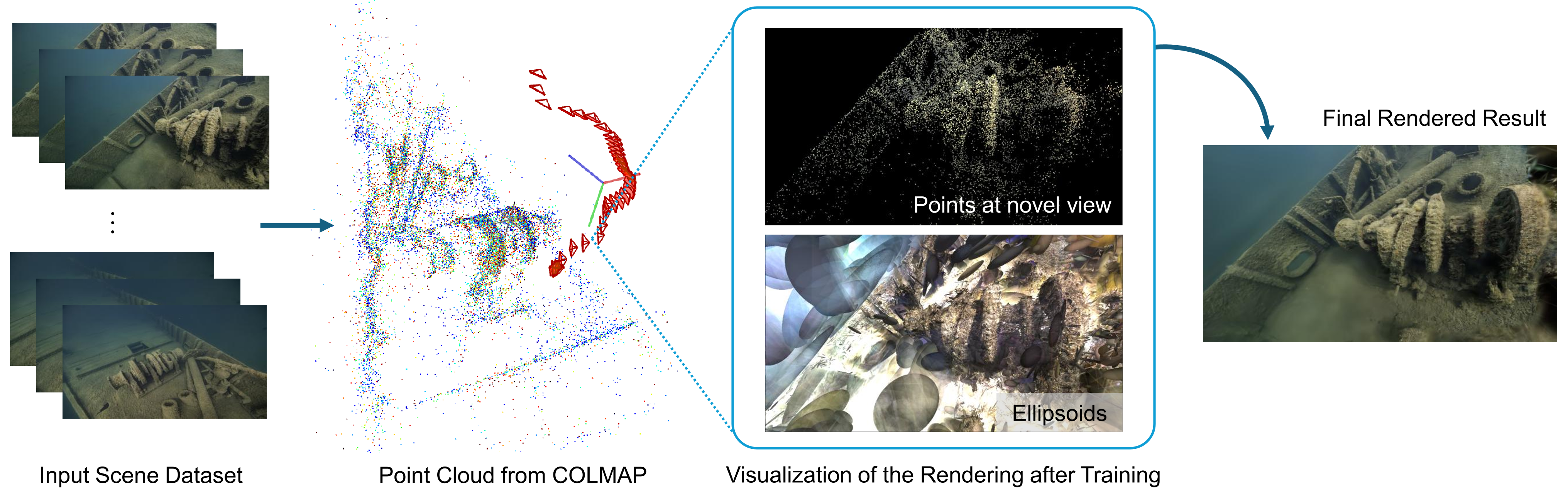}
    \caption{Workflow showing the point cloud created from input images using COLMAP, which is noisy and sparse in some areas. After training, the point cloud used for novel view rendering becomes cleaner. The ellipsoids of each 3D point are shown, illustrating how they are used to create the final rendered result.}
    \label{fig:pointclound}
\end{figure*}

% -----------------------------------------------------

\subsection{Quantitative Comparison}

 Quantitative performance is evaluated using three standard metrics: Peak Signal-to-Noise Ratio (PSNR), Structural Similarity Index Measure (SSIM), and Learned Perceptual Image Patch Similarity (LPIPS)~\citep{8578166}, which measure pixel-level accuracy, structural similarity, and perceptual quality, respectively.

\autoref{tab:comparison} presents a quantitative comparison of the proposed method against six baseline approaches across three datasets. Averaged across all datasets, the proposed method achieves the best performance on all three metrics, demonstrating consistent improvements in both reconstruction accuracy and perceptual quality.

On the SeaThru-NeRF dataset, our method outperforms the strongest baseline, RUSplatting, with an average gain of 0.67 dB in PSNR, a 1.05\% increase in SSIM, and a 5.70\% reduction in LPIPS. Notably, improvements are observed across all scenes, indicating robust performance under varying water conditions.

On the more challenging Submerged3D dataset, which features sparse viewpoints and severe image degradation, the proposed method continues to outperform existing approaches. Compared to UW-GS, it achieves average improvements of 1.80 dB in PSNR and 3.72\% in SSIM, along with a 1.13\% reduction in LPIPS. These gains demonstrate the effectiveness of the proposed approach in handling low-visibility and sparse-view scenarios.

On the S-UW dataset across four scenes, our method achieves the best average performance in terms of PSNR (25.6425 dB), along with competitive SSIM (0.8121) and LPIPS (0.2017). The strong PSNR results likely stem from Stage 2 of our optimization strategy, where the $\ell_2$ term is emphasized. It is worth noting that different methods exhibit distinct trade-offs. For instance, WaterSplatting achieves the lowest average LPIPS, suggesting stronger perceptual similarity, while RUSplatting attains high SSIM and strong PSNR in certain scenes. In contrast, our method provides more balanced performance across all metrics and scenes, without significant degradation in any particular setting. This consistency is particularly important for shallow-water scenarios in S-UW, where flickering and non-rigid motion introduce additional challenges.

Over all, these results demonstrate the effectiveness of our semantic-guided, stage-wise optimization and Gaussian reallocation strategies in enhancing both structural consistency and perceptual quality under challenging underwater conditions, from relatively controlled reef environments to highly degraded deep-sea scenes.

% -===========================================================
\begin{table}[!t]
    \centering
        \caption{Ablation results averaged over all scenes from the SeaThru-NeRF, Submerged3D and S-UW datasets. Red, orange, and yellow denote the best, second-best, and third-best results, respectively.}
   % \resizebox{0.99\linewidth}{!}{
   \begin{tabular}{r|cccccc}
        \toprule
        Variant & SG & 
        SO   & 
        AR & PSNR$\uparrow$ & SSIM$\uparrow$ & LPIPS$\downarrow$ \\
        \midrule
        M1 &  \xmark& \cmark& \cmark&27.4453 & 0.8373 & 0.2426 \\
        M2 & \cmark& \xmark& \cmark&\cellcolor{lightyellow}27.8131 & \cellcolor{lightyellow}0.8511 & \cellcolor{lightyellow}0.2355 \\
        M3 & \cmark& \cmark& \xmark& \cellcolor{lightorange}27.8604 & \cellcolor{lightorange}0.8632 & \cellcolor{lightorange}0.2297 \\
        Ours &  \cmark& \cmark& \cmark&\cellcolor{lightred}28.0508 & \cellcolor{lightred}0.8636 & \cellcolor{lightred}0.2272 \\
        \bottomrule
    \end{tabular} \\
 %   \vspace{1pt}
\begin{flushleft} 
SG: Semantic-guided Gaussians; SO: Stage-wise optimization;\\
AR: Adaptive Gaussian primitives reallocation
\end{flushleft}
    \label{tab:ablation}
\end{table}

\subsection{Qualitative Comparison}

The qualitative results in \autoref{overall_results} demonstrate the visual performance of the proposed method in comparison with existing approaches. Baseline methods exhibit noticeable limitations in reconstructing fine geometric structures and maintaining background consistency, particularly in regions affected by scattering and low visibility (in the yellow and red highlighted regions in the first row of the figure, which is from the seaThru-NeRF dataset). For example, several methods fail to recover the fine, spiny structures of coral or produce inconsistent background textures. In contrast, the proposed method preserves these intricate details, resulting in sharper and more faithful reconstructions.

For the Submerged3D dataset, the second row of \autoref{overall_results} shows that the proposed method achieves improved geometric accuracy and more consistent color representation, whereas competing methods exhibit visible artifacts (e.g., RUSplatting) or blurred structures (e.g., UW-GS). The third row further confirms the advantage of the proposed method in maintaining fine structural details and perceptual quality across challenging scenes. The fourth row shows the results on the S-UW dataset. While the quantitative results in \autoref{tab:comparison} show that RUSplatting achieves the best PSNR, the qualitative results indicate that it struggles to reconstruct content accurately when lighting varies across views. Although WaterSplatting achieves the best perceptual quality metric (LPIPS), its results, while appearing sharper than those of our proposed SWAGSplatting, also introduce artifacts, as shown in the cropped regions. Overall, the results indicate that the proposed approach produces more consistent and visually coherent reconstructions across diverse underwater conditions.

\autoref{fig:pointclound} illustrates the 3D rendering workflow. The figure shows that the COLMAP reconstruction contains noticeable noise and irregular point distributions, particularly in regions affected by low visibility and scattering. These artifacts arise from unreliable feature matching and inaccurate depth estimation under challenging underwater conditions. In contrast, the point cloud produced by our method is more coherent and spatially consistent after optimization. Noise is substantially reduced, and structural details are better preserved, resulting in a cleaner and more complete scene representation. These results indicate that the proposed method yields higher-quality geometric representations than conventional photogrammetry-based approaches in degraded underwater environments, where additional post-processing such as filtering out outliers and noise, and optionally downsampling or cleaning the data, is often required to obtain usable reconstructions.

\begin{figure}
    \centering
    \includegraphics[width=\linewidth]{ 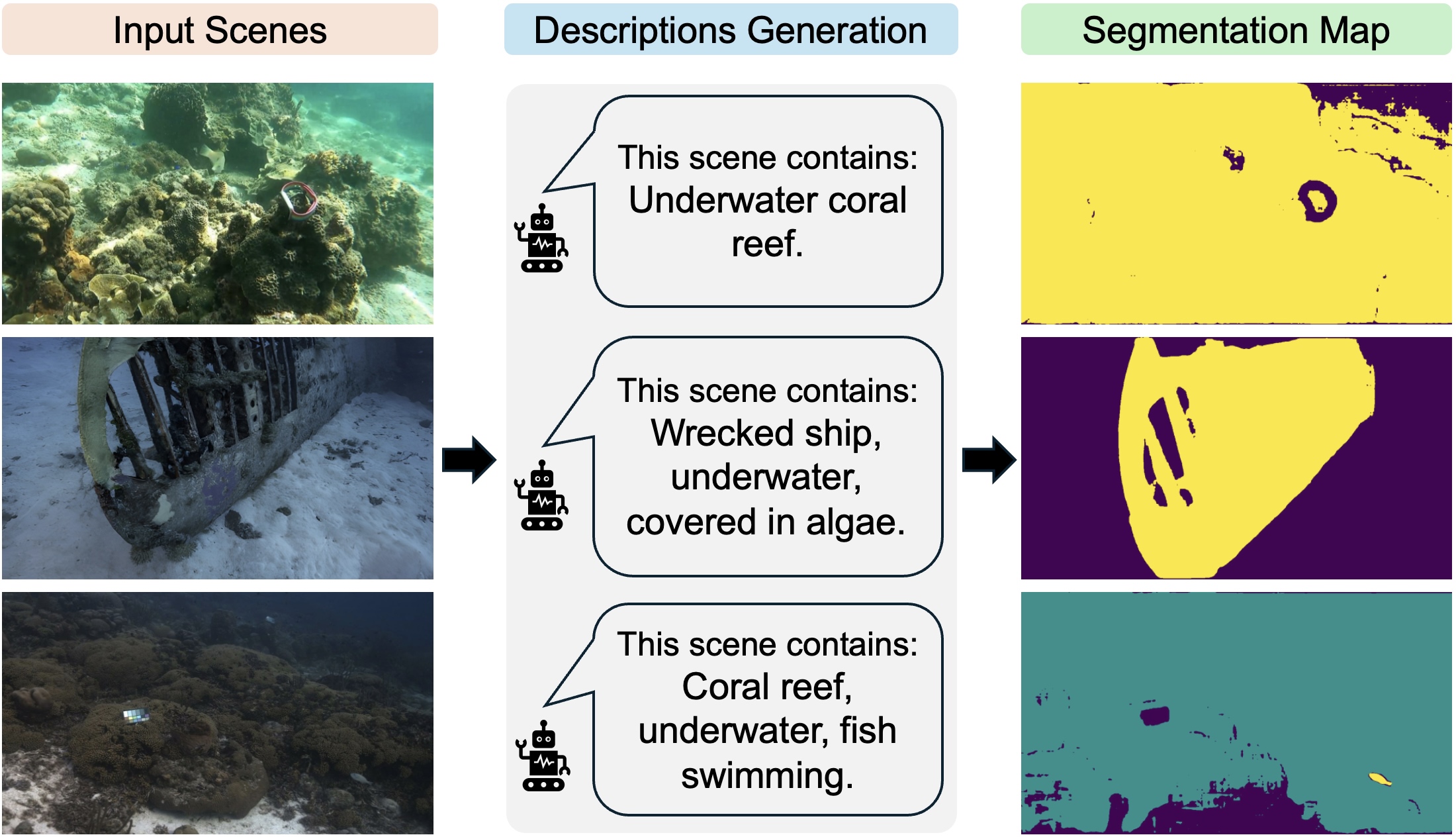}
    \caption{Examples of semantic guidance. From left to right: input scenes, generated descriptions, and segmentation maps.}
    \label{seg_figure}
\end{figure}
% -----------------------------------
\subsection{Ablation Study}

To assess the contribution of each component in the proposed framework, we conduct ablation experiments, the configurations and results of which are summarized in \autoref{tab:ablation}. Starting from the full model, removing any individual component leads to a consistent degradation in performance across all metrics, indicating that each module contributes positively to the overall reconstruction quality. %In particular, variants M1, M2, and M3 exhibit lower PSNR and SSIM, and higher LPIPS, compared to the full model.

Among these, removing semantic guidance (M1) results in the largest performance drop. This highlights the role of the semantic module as a perceptual-level regularizer, improving region-level consistency without relying on explicit category labels. By leveraging CLIP embeddings, semantically similar regions are encouraged to exhibit coherent representations, thereby stabilizing reconstruction in areas with ambiguous or degraded photometric cues. \autoref{seg_figure} shows examples of generated descriptions and their corresponding semantic segmentation maps.
Given the limited availability of annotated underwater datasets, evaluation is conducted on established benchmarks, and conclusions are therefore restricted to reconstruction performance under these conditions.

Variant M2 removes the stage-wise optimization component while retaining both semantic guidance and Gaussian reallocation. Without staged training, the model jointly optimizes geometry and appearance throughout the entire training process, which can lead to instability and suboptimal refinement, particularly under degraded imaging conditions. The stage-wise strategy in the full model helps decouple global structure learning from fine-grained appearance optimization, enabling more stable convergence and improved reconstruction fidelity.

Removing the adaptive Gaussian reallocation component (M3) results in a comparatively smaller performance decrease. This is expected, as the reallocation mechanism primarily improves representation efficiency by redistributing primitives from low-contribution regions to areas with higher reconstruction error. Its impact is more localized (as shown in \autoref{fig:Reallocation}) and therefore less pronounced in scene-level average metrics under a fixed primitive budget.

Overall, the full model achieves the best performance, indicating that the proposed components provide complementary benefits when combined.

% ========================================
\section{Conclusions}
This paper presents a semantic-guided 3D Gaussian Splatting framework for underwater scene reconstruction. The proposed method extends each Gaussian primitive with a learnable semantic feature, supervised using CLIP-based embeddings to promote semantic–geometric consistency. A semantic loss is introduced to encourage the preservation of high-level structural relationships, leading to improved perceptual quality.
In addition, a stage-wise optimization strategy is employed to enhance training stability, and an adaptive Gaussian reallocation mechanism is proposed to improve representation efficiency by redistributing primitives within a fixed budget. Together, these components enable more consistent and detailed reconstructions under challenging underwater conditions. Experimental results demonstrate that the proposed method achieves improved performance across multiple benchmarks. The code will be released upon acceptance.

\printcredits

\section*{Declaration of competing interest}
The authors declare that they have no known competing financial interests or personal relationships that could have appeared to influence the work reported in this paper.

\section*{Acknowledgments}
This work was supported by the UKRI MyWorld Strength in Places Programme (SIPF00006/1) and EPSRC ECR (EP/Y002490/1).

\bibliographystyle{cas-model2-names}

\bibliography{cas-refs}



\end{document}